\documentclass[a4paper,fleqn]{cas-dc}

\usepackage{placeins}
\usepackage{xcolor}
\usepackage{rotating}
\usepackage[colorlinks]{hyperref}
\usepackage{colortbl}
  \newcommand{\myrowcolour}{\rowcolor[gray]{0.925}}
\usepackage[authoryear,longnamesfirst]{natbib}

\begin{document}
\let\WriteBookmarks\relax
\def\floatpagepagefraction{1}
\def\textpagefraction{.001}
\shorttitle{}
\shortauthors{Rownak Ara Rasul et al.}

\title [mode = title]{An Evaluation of Machine Learning Approaches for Early Diagnosis of Autism Spectrum Disorder}

\author[First]{Rownak Ara Rasul} 
\ead{rownakararasul30@gmail.com}

\author[Second]{Promy Saha}
\ead{promysaha95@gmail.com}

\author[Third]{Diponkor Bala} \corref{cor1}
\ead{diponkor@cityuniversity.ac.bd}

\author[Fourth]{S M Rakib Ul Karim} 
\ead{skarim@mail.missouri.edu}

\author[First]{Md. Ibrahim Abdullah}
\ead{ibrahim@cse.iu.ac.bd}

\author[Fifth]{Bishwajit Saha}
\ead{sahab@rpi.edu}

\cortext[cor1]{Corresponding author.}

\address[First]{Department of Computer Science and Engineering, Islamic University, Kushtia-7003, Bangladesh}
\address[Second]{Khulna Medical College \& Hospital, Khulna, Bangladesh}
\address[Third]{Department of Computer Science and Engineering, City University, Savar, Bangladesh}
\address[Fourth]{Department of Electrical and Computer Engineering, University of Missouri, Columbia, Missouri, USA}
\address[Fifth]{Department of Computer Science, Rensselaer Polytechnic Institute, Troy, NY, USA}

\begin{abstract}
Autistic Spectrum Disorder (ASD) is a neurological disease characterized by difficulties with social interaction, communication, and repetitive activities. While its primary origin lies in genetics, early detection is crucial, and leveraging machine learning offers a promising avenue for a faster and more cost-effective diagnosis. This study employs diverse machine learning methods to identify crucial ASD traits, aiming to enhance and automate the diagnostic process. We study eight state-of-the-art classification models to determine their effectiveness in ASD detection. We evaluate the models using accuracy, precision, recall, specificity, F1-score, area under the curve (AUC), kappa, and log loss metrics to find the best classifier for these binary datasets. Among all the classification models, for the children dataset, the SVM and LR models achieve the highest accuracy of 100\% and for the adult dataset, the LR model produces the highest accuracy of 97.14\%. Our proposed ANN model provides the highest accuracy of 94.24\% for the new combined dataset when hyperparameters are precisely tuned for each model. As almost all classification models achieve high accuracy which utilize true labels, we become interested in delving into five popular clustering algorithms to understand model behavior in scenarios without true labels. We calculate Normalized Mutual Information (NMI), Adjusted Rand Index (ARI), and Silhouette Coefficient (SC) metrics to select the best clustering models. Our evaluation finds that spectral clustering outperforms all other benchmarking clustering models in terms of NMI and ARI metrics while demonstrating comparability to the optimal SC achieved by k-means. The implemented code is available at  \href{https://github.com/diponkor-bala/Autism-Spectrum-Disorder}{GitHub}.
 
\end{abstract}


\maketitle

\section{Introduction}
Autism spectrum disorder (ASD) is a neurodevelopmental illness that affects 1 in 68 children and 1 in 88 adults. Individuals with ASD have trouble in various domains, including social interactions, communication, play, and stereotypical activities~\citep{de2004autism}. These challenges often manifest as a tendency to engage in repetitive and highly restricted routines.
Early ASD diagnosis may reduce developmental, behavioral, and other repetitive social activities. Traditional ASD diagnosis methods include the Autism Diagnostic Interview (ADI)~\citep{rutter2003autism} and Autism Diagnostic Observation Schedule-Generic (ADOS)~\citep{lord2000autism}, which involve observational and behavioral tests conducted by physicians and parents. The AODS includes a number of subtests to measure specific aspects of day-to-day life, such as social interaction, attitude toward games and sports, and creative use of everyday materials. This process of diagnosis and detection of ASD patients is laborious, time-consuming, and inefficient.  In some cases, the average time for this type of diagnosis and detection takes more than three years~\citep{landa2008diagnosis}. To avoid these delays, researchers have come up with self-screening and parent-screening methods that give a first idea of who might have ASD traits. After the parents have noticed ASD traits, the patient needs to be tested by a certified, well-trained clinician and professional doctor to find out if the patient has ASD traits or not. This used to take up to 90 minutes in the past. This process additionally requires an extensive number of both professionally skilled and unskilled laborers, which is expensive and time-consuming. Apart from this, the longer period of determining ASD, administers additional disadvantages to the patient's daily life, which delays their treatment, speech and behavioral therapies, and other medication that might improve the quality and socio-cultural activities of the ASD patients. 

Using artificial intelligence (AI)~\citep{shahamiri2020autism} and machine learning (ML)~\citep{song2015decision},~\citep{hsu2003practical} approaches, a large amount of research was conducted to develop an effective and efficient method to correctly identify ASD patients with commendable ASD detection accuracy and efficiency. It was done to reduce all the disadvantages, including the longer time of diagnosis, its expense, and the additional amount of manpower. However, there is a shortage of substantial research that thoroughly investigates the benefits and drawbacks of using different aspects of machine learning algorithms for the reliable and effective detection of ASD symptoms. 

In this study, we attempted to close this knowledge gap by discussing several aspects of using machine learning approaches to ASD diagnosis. In this work, we tried to bridge this gap by addressing different aspects of the application of machine learning techniques in the diagnosis of ASD patients by applying both supervised and unsupervised approach.For the supervised technique, we have applied eight machine learning algorithms with eight different performance metrics for the evaluation of the model's performance. Moreover, to the best of our knowledge, there is no work that applies any clustering algorithm to these ASD datasets to see how machine learning models work when true labels are not provided. We first propose and apply different types of popular clustering algorithms on these datasets to see how they perform in terms of Silhouette Coefficient~\citep{rousseeuw1987silhouettes}, Normalized Mutual Information~\citep{vinh2009information} \& Adjusted Rand Index~\citep{hubert1985comparing} scores. We apply K-means~\citep{macqueen1967classification}, Spectral Clustering~\citep{donath1973lower}, Hierarchicla/Agglomerative Clustering~\citep{johnson1967hierarchical}, GMM~\citep{dempster1977maximum} and Birch~\citep{zhang1996birch} on these ASD datasets. 

Our major contributions to this study are listed as follows:
\begin{itemize}
  \item We have studied eight classification models to see which model works best to identify ASD, and we also studied five popular clustering methods to see how machine learning models work when true labels are not provided for these ASD datasets.
    
  \item We combine all the datasets to make a dataset of the whole population to get a holistic idea of how ASD patients prevail and how well our different types of ML algorithms perform in comparison to segmented datasets.
  
  \item We have calculated the best features of each category of the datasets, which has helped us boost the performance of the ML models.

  \item We conducted a rigorous hyper-parameter search approach to get the best parameters for each of the models to get the optimum results.

  \item Finally, we have developed a Graphical User Interface (GUI) where we have integrated our models, which will help clinicians early detect ASD.

\end{itemize}

The remaining portion of the text is organized as follows: Related research is discussed in Section 2, and datasets and the classification and clustering methods utilized in our study are discussed in Section 3. In Section 4 we describe the main aspects of the assessment measures we have utilized for the performance study. Results and discussion of this effort are presented in Section 5, and a conclusion is formed in Section 6.

\section{Related Works}
The standard of living for individuals with ASD as well as their families may be significantly enhanced with early identification and diagnosis, which allows for prompt intervention and support. Because of this, several research articles have been published about how machine learning can be used to implement screening solutions. In recent years, machine learning technologies have gained popularity in the area of ASD detection and diagnosis due to their capacity to evaluate large datasets, spot patterns, and provide reliable predictions.

For examples,  
~\cite{thabtah2019early} proposed modifying the current screening tool DSM-5 (Diagnostic and Statistical Manual of Mental Disorders, 5th edition) to detect ASD disease and incorporating machine learning for more rapid and accurate ASD recognition. To expand on their ideas,  ~\cite{thabtah2020new} suggested a machine learning method called Rules-Machine Learning (RML) to quickly and correctly identify ASD. They used a set of data from a mobile app called ASDTests[6] to compare their method with RIPPER (Repeated Incremental Pruning to Produce Error Reduction), AdaBoost (Adaptive Boosting), RIDOR (Ripple Down Rule Learner), Nnge (Non-Nested Generalization), Bagging, CART (Classification and Regression Tree), and C4.5 (used in Data Mining as a Decision Tree Classifier) classifiers. Their method did better than all of them except for C4.5.
~\cite{cook2019towards} used machine learning methods to look at how children with ASD and usually developing children spoke and how they used language. The study showed that it might be possible to classify ASD cases correctly. However, the success of the system was found to be sensitive to changes in the recording surroundings and to different languages. Because of this, it was hard to use in real-world situations. Using the SHAP (SHapley Additive exPlanations) approach ~\cite{bala2022efficient} boosted accuracy and ranked features for analysis within different subsets. They introduced a machine-learning model for ASD across age groups and utilized a variety of approaches and classifiers. SVM demonstrated superior performance and gained high accuracies for toddlers (97.82\%), children (99.61\%), adolescents (95.87\%), and adults (96.82\%). ~\cite{gaspar2022optimized} focused on using machine learning and eye-tracking technology for early detection. The researcher introduced a new method using Kernel Extreme Learning Machine (KELM), objective gaze tracking data, feature extraction, and data augmentation. Implementing the Giza Pyramids Construction (GPC) method optimized the model and achieved 98.8\% accuracy in classifying ASD cases.

~\cite{nishat2022detection} developed an ML-based model aiming to predict ASD and associated psychological disorders impacting social behavior. They applied quadratic discriminant and linear analysis on UCI reservoir data to detect, analyze, and plan treatment for ASD. With hyperparameter adjustments, the Quadruple Analysis Algorithm (QDA) achieved an impressive 99.77\% accuracy, affirming the model's effectiveness. ~\cite{hyde2019applications} introduced a concise web-based survey to detect parental autism with 15 questions. Using supervised machine learning, particularly SVM, they evaluated ASD-related text data. The collection of a large dataset was difficult, although the supervised machine learning (SVM) algorithms performed admirably with a high degree of accuracy. Using data from the Dhaka Shishu Children's Hospital,~\cite{tariq2019detecting} identified developmental gaps, including speech and language issues, utilizing data from Bangladesh and the United States. Experimenting with various train-test splits, they achieved a 75\% accuracy by training classifiers on clinical scoresheets rather than live video data features.
 ~\cite{rouhi2019emotify} presented a spoken educational game using machine learning that helps ASD children how to detect and express emotions. The game emphasizes four emotional states: happiness, sorrow, rage, and neutrality. He relied on the EMOTIFY (Emotional game for children with autism spectrum disorder): Website application. Child learning and user skill testing are two parts. Voice with human facial expressions data is not cross-validated here. ~\cite{sharif2022novel} employed various machine learning algorithms, including Linear Discriminant Analysis, Random Forest, SVM, K-Nearest Neighbors (KNN), and Multi-layer Perceptron, to detect ASD using the ABIDE-I dataset. Their results indicated accuracies ranging from 55\% to 65\% while ~\cite{sherkatghanad2020automated} utilized a Convolutional Neural Network (CNN) model on the same dataset and achieved a higher accuracy of 70.22\%. 
 Another study, ~\cite{usta2019use} intended to test machine learning methods on big datasets to determine outcome factors. The 254 baseline form components were used to assess four machine learning methods: Naive Bayes, Generalized Linear Model (GLM), Logistic Regression, and Decision Tree. They got the best AUC and accuracy using a decision tree (AUC = 70.7\%, sensitivity = 81.1\%, specificity = 61.3\%). The research is limited by the absence of complicated neurological comorbidity studies.
In a recent study, ~\cite{wei2023early} employed machine learning models to provide a more reliable and interpretable technique for distinguishing early children ASD, DLD (Developmental Language Disorder), and GDD (Global Developmental Delay). The best-performing model, eXtreme gradient boosting (XGB), obtained an accuracy of 78.3\% on an external dataset by incorporating data from several behavioral and developmental tests.
According to ~\cite{cavus2021systematic}, a systematic approach was used to enhance the Grasshopper Optimization Algorithm by including the Random Forest (RF) classifier. Traditional ASD disease recognition systems are time-consuming and may be misleading, according to the authors, ~\cite{yin2021diagnosis} of this study. They applied a deep learning-based approach which was presented to detect neurological disease patients, achieving 79.2\% accuracy on brain functional MRI (Magnetic Resonance Imaging) data using a deep neural network auto-encoder.
~\cite{heinsfeld2018identification} proposed a deep learning auto-encoder model with cross-validation folds technique to detect ASD disease and used a dataset called ABIDE (Autism Brain Imaging Data Exchange), which is publicly available in online. Their model tried to find a functional connectivity pattern that identified ASD patients' brain imaging data.
Similarly, ~\cite{kashef2022ecnn} experimented with deep learning methods to detect autism spectrum disorder (ASD) in brain imaging data from the ABIDE database. They assessed brain connection patterns using CNN and obtained up to 80\% accuracy. The results emphasized impaired brain connection as a significant biomarker of autism spectrum disorder.
 
The other study, ~\cite{mashudi2021classification}, used a simulated environment called Waikato Environment for Knowledge Analysis (WEKA) to look for ASD in a group of 703 patients and people who were not patients. The people had 16 different traits. The study used SVM, J48 (decision tree algorithm), KNN, Naïve Bayes, Bagging, AdaBoost, and Stacking, applying 3, 5, and 10-fold cross-validation with 100\% accuracy for Naïve Bayes, Bagging, SVM, Stacking, and J48 methods. A study by ~\cite{abdullah2019evaluation} looked at how to find ASD early on. They used the Autism Spectral Questionnaires (AQ) dataset to test KNN with k-fold cross-validation, Random Forest, and Logistic Regression.  Using Chi-square and LESSO (Least Absolute Shrinkage and Selection Operator) for feature selection, Logistic Regression reached 97.41\% accuracy, the highest among the classifiers. ~\cite{qureshi2023prediction} compared ASD prediction based on application type, simulation method, comparison methodology, and input data, aiming to offer a unified framework for researchers. With an accuracy of 89.23\%, the random forest outperformed the other methods.
~\cite{alteneiji2020autism} separated the population into infants, toddlers, and adolescents in order to apply existing machine-learning techniques. For the adolescent group of ASD populations, neural networks provided an ASD recognition accuracy of 99.04\%. ~\cite{ahmed2022toward} presented an 82-question ASD symptom questionnaire, which resulted in a dataset. An artificial neural network performed here with an 89.8\% of accuracy. To diagnose and prioritize autism spectrum conditions, Artificial Intelligence (AI) is investigated. ~\cite{joudar2022triage} evaluated scholarly material from 2017 to 2022 in three categories: diagnosing ASD, prioritizing dangerous genes, and telehealth triage. The article emphasized the challenges and gaps in current research and suggested a new methodology for triaging and prioritizing ASD patients based on severity using AI.

Our study prioritizes early diagnosis of ASD in children and adults analyzing two separate datasets encompassing information on both age groups while proposing a unified combined dataset. Employing various machine-learning methodologies and leveraging cross-validation, we aim to predict ASD among different age groups.

\section{Materials \& Proposed Methods}
\begin{figure*}
\centering
  \includegraphics[width=\linewidth]{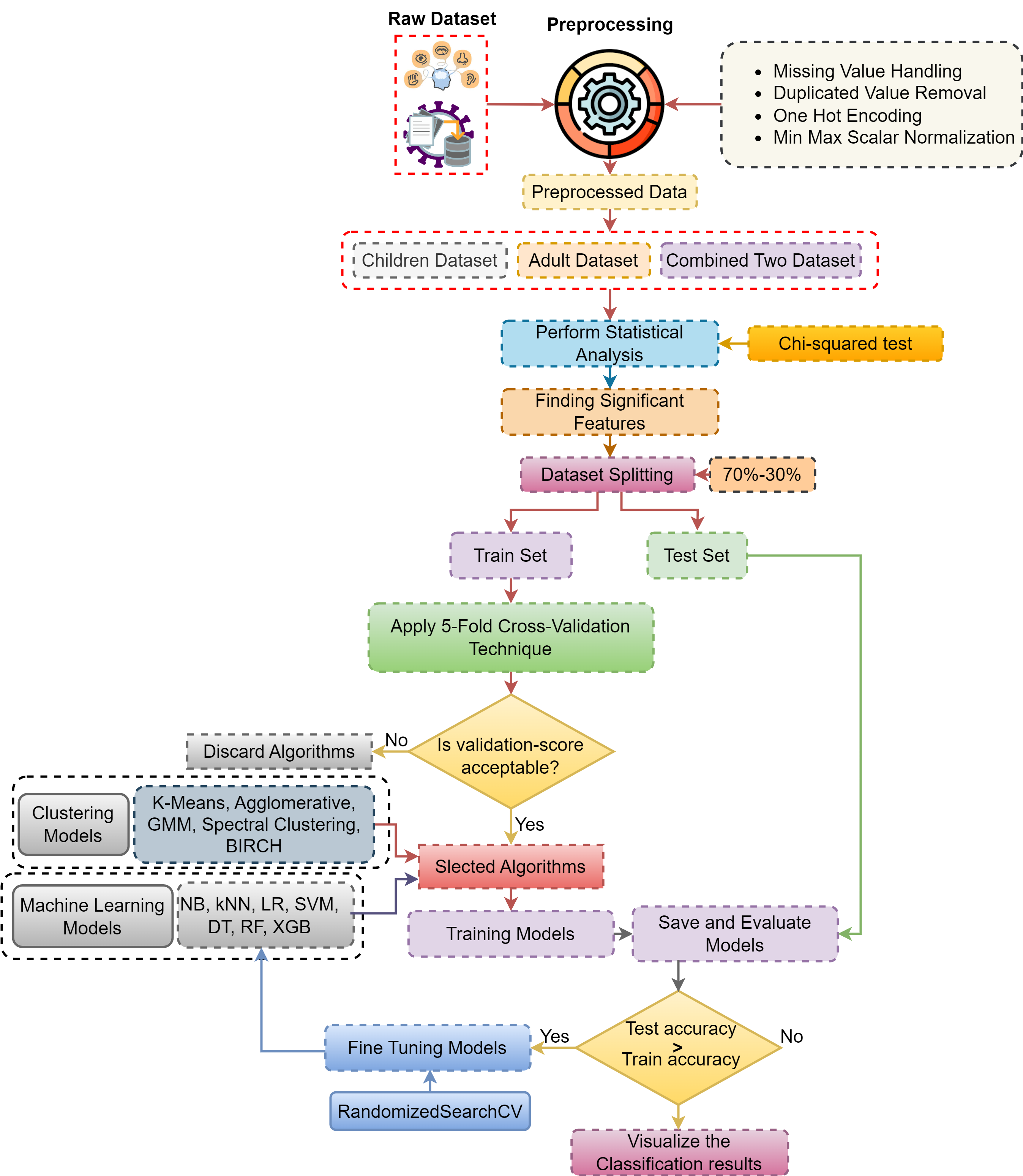}
  \caption{The schematic diagram of the proposed ASD detection system}
  \label{fig:flowchart}
\end{figure*}

Our research mainly focuses on the fact that ASD in children and adults should be diagnosed at an early stage. In this study, we used two raw public datasets comprised of different features of children and adults. We also introduced a third dataset that comes from the combination of these two datasets. Next, in the preprocessing stages, missing value handling, removing the duplicate values, and finding the significant features by using the chi-square statistical method are done, and then the dataset splitting steps have been completed. Then we applied a 5-fold cross-validation technique to select the machine learning algorithms that were used to perform the classification tasks later. After selecting the ML algorithms, we trained the algorithms on the training dataset and then evaluated them on the test dataset. After the model evaluation, the classification results are visualized. Here's the schematic diagram of the proposed methodology, shown in Figure \ref{fig:flowchart}, and how each step of the proposed method is explained in more detail.

\subsection{Data Collection}
\begin{figure*}[]
\centering

    \includegraphics[width=.40\textwidth]{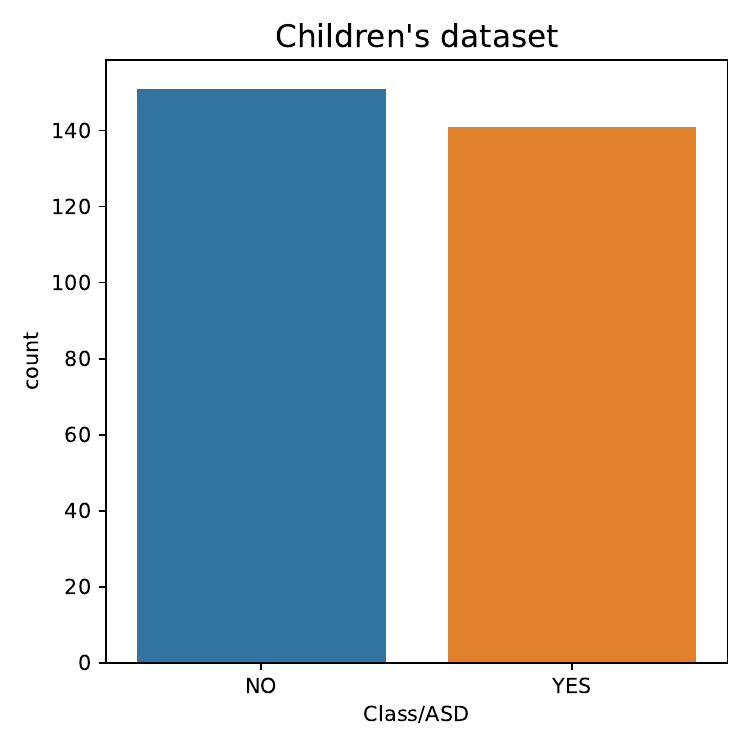}
    \includegraphics[width=.40\textwidth]{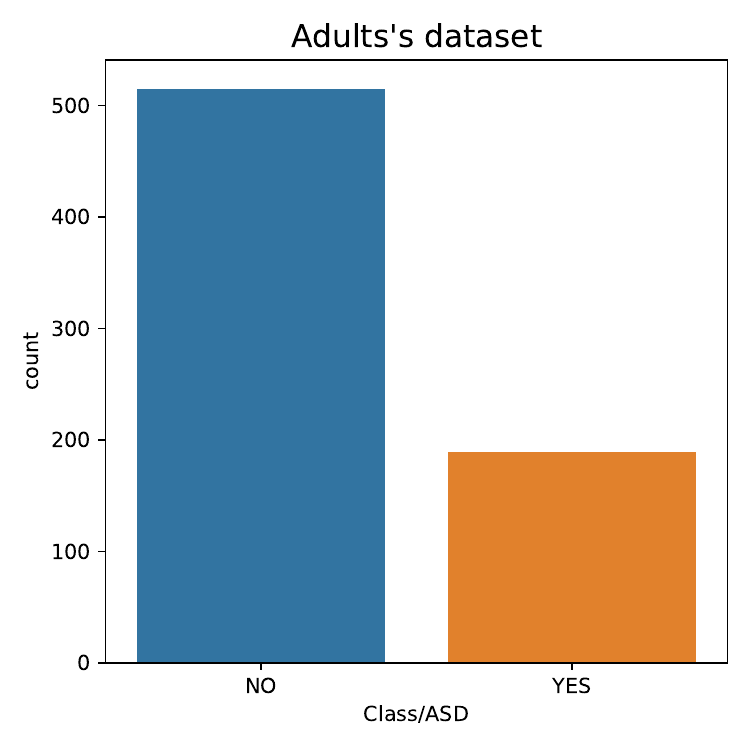}

\caption{Data distribution for Children and Adult dataset}
\label{fig:distribution}
\end{figure*}

The global incidence of Autism Spectrum Disorder (ASD) is becoming more common throughout the world, yet there are very few publicly available datasets that are specifically devoted to studying the disorder. At present, genetics-focused databases predominate, and there is a shortage of clinical screening datasets for autism. The ASD in Children’s\footnote{ https://doi.org/10.24432/C5659W}~\citep{misc_autistic_spectrum_disorder_screening_data_for_children___419} and Adult’s\footnote{https://doi.org/10.24432/C5F019}~\citep{misc_autism_screening_adult_426} datasets used in this study were obtained from a publicly accessible UCI Repository, where the Children dataset contains 292 instances and 21 attributes, and the Adult dataset contains 704 instances and 21 attributes. There are 292 observations in this children's dataset, ranging in age from 4 to 11. Both their unique traits and responses to the ten questions (AQ10) are documented. Every response to the query has a code of either 0 or 1. In AQ10, a score of 0 is the lowest possible, with a maximum of 10. The data also includes the final score. Class (ASD) is our response variable. The detailed summarizing of all three datasets are given in the Table \ref{tab:distibution} and the pictorial representation of the dataset's individual class instance distribution is shown in Figure \ref{fig:distribution}.

\begin{table*}[t]
\caption{Types of Dataset with Instances and Attributes.
\vspace{-0.15in}}
\label{tab:distibution}
\begin{center}
{\footnotesize
\begin{tabular}{ccccccc}
\hline
\multirow{2}{*}{Sr. No.} & \multirow{2}{*}{Dataset Name}   & \multirow{2}{*}{Attribute Type}    & \multirow{2}{*}{Number of Attributes} & \multicolumn{3}{l}{Number of Instances}                    \\ \cline{5-7} 
                         &                                 &                                    &                                       & \multicolumn{1}{l}{Yes} & \multicolumn{1}{l}{No}  & Total \\ \hline
01                       & ASD Screening data for Children & Categorical, binary and continuous & 21                                    & \multicolumn{1}{l}{141} & \multicolumn{1}{l}{151} & 292   \\ \hline
02                       & ASD Screening data for Adult    & Categorical, binary and continuous & 21                                    & \multicolumn{1}{l}{189} & \multicolumn{1}{l}{515} & 704   \\ \hline
03                       & ASD Screening Combined data     & Categorical, binary and continuous & 21                                    & \multicolumn{1}{l}{330} & \multicolumn{1}{l}{666} & 996   \\ \hline
\end{tabular}
}
\end{center}
\end{table*}

\begin{table*}[t]
\caption{Feature descriptions}
\centering
 \begin{tabular}{ p{2.5cm} p{3cm} p{10cm}}
 \Xhline{2\arrayrulewidth}
  \myrowcolour
 Feature No. & Attributes & Description\\
 \hline
 01  & age & age of an individual in years.\\
  \myrowcolour
 02 & gender  & gender of an individual.\\
 
 03 & ethnicity & belonging of an individual to a social group.\\
  \myrowcolour
 04 & jaundice & whether the individual was having jaundice at birth or not.\\
 
 05 & autism & an immediate family member has a pervasive developmental disorder.\\
  \myrowcolour
 06 & relation & relation with the suspected individual.\\
 
 07 & country\_of\_res & country of residence.\\
  \myrowcolour
 08 & used\_app\_before & any prior screening or test for ASD.\\
 
 09 & age\_desc & Age category.\\
  \myrowcolour
 10 & result & Screening result score.\\
 
 \Xhline{2\arrayrulewidth}
 
 \end{tabular}
 \label{table:feature1_tbl}
\end{table*}

\begin{table*}
\caption{A1-A10 scores: 1(YES)/ 0(NO) based on the question asked in screening.}
\centering
 \begin{tabular}{ p{2.5cm} p{3cm} p{10cm}}
 \Xhline{\arrayrulewidth}
  \myrowcolour
 Feature No. & Attributes & Description\\
 \hline
 10  & A1 score & The answer code of: Does the person speak very little and give unrelated answers to questions?\\
  \myrowcolour
 11 & A2 score  & The answer code of: Does the person not respond to their name or avoid eye contact?\\
 
 12 & A3 score & The answer code of: Does the person not engage in games of pretend with other children?\\
  \myrowcolour
 13 & A4 score & The answer code of: Does the person struggle to understand other people’s feelings?\\
 
 14 & A5 score & The answer code of: Is the person easily upset by small changes?\\
  \myrowcolour
 15 & A6 score & The answer code of: Does the person have obsessive interests?\\
 
 16 & A7 score & The answer code of: Is the person over or under-sensitive to smells, tastes, or touch?\\
  \myrowcolour
 17 & A8 score & The answer code of: Does the person struggle to socialize with other children?\\
 
 18 & A9 score & The answer code of: Does the person avoid physical contact?\\
  \myrowcolour
 19 & A10 score & The answer code of: Does the person show little awareness of dangerous situations?\\
 
 \Xhline{2\arrayrulewidth}
 \end{tabular}
 \label{table:feature2_tbl}
\end{table*}

\begin{figure*}
\centering
  \includegraphics[width=0.85\linewidth]{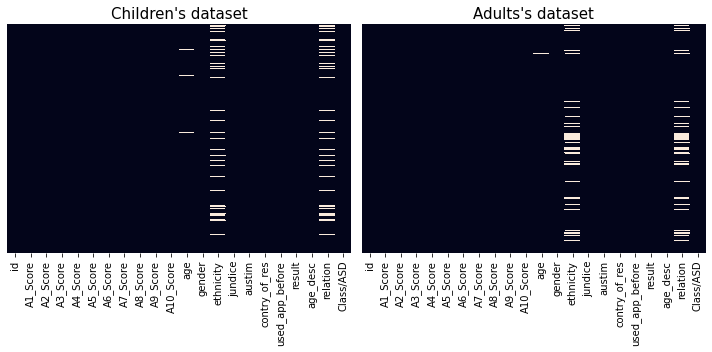}
  \caption{Visualization of the missing values using the seaborn heatmap.}
  \label{fig:missing_value}
\end{figure*}

\begin{figure*}[]
\centering

    \includegraphics[width=.49\textwidth]{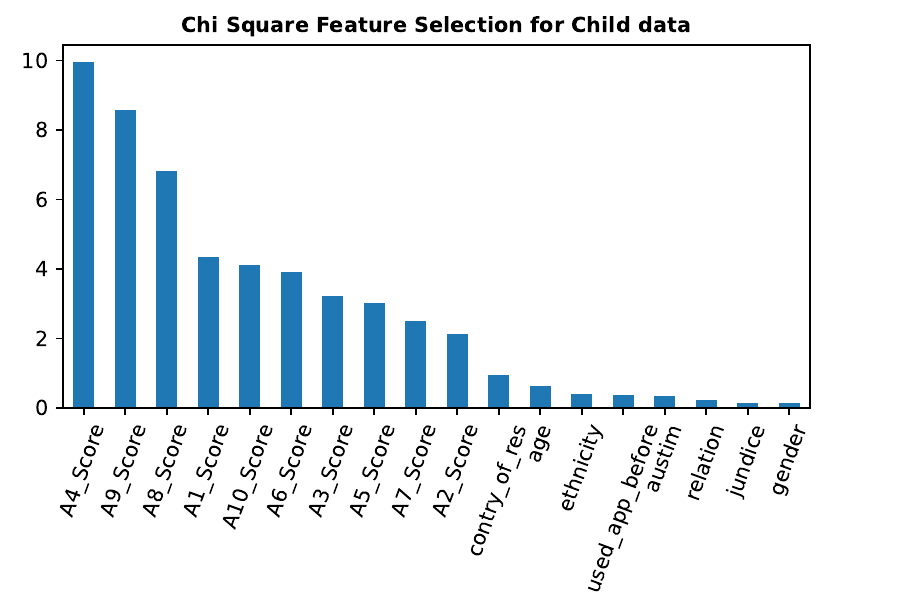}
    \includegraphics[width=.49\textwidth]{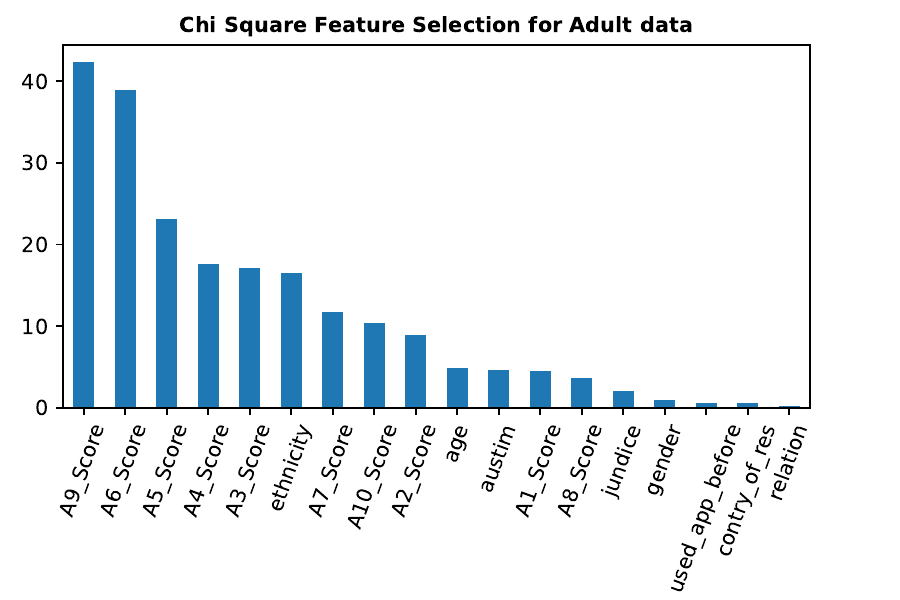}
    \includegraphics[width=.49\textwidth]{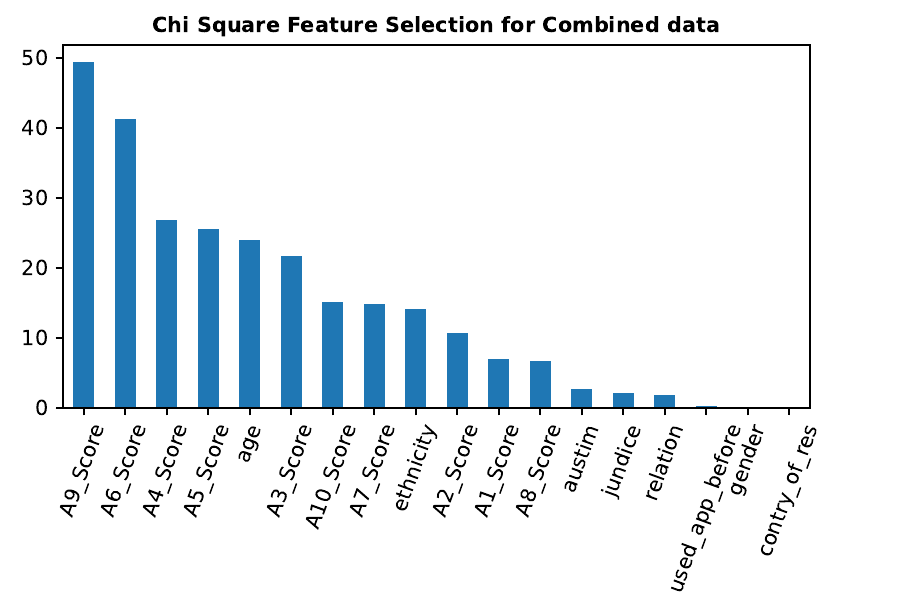}

\caption{Feature rank according to Chi-Square test for Children, Adult and Combined dataset}
\label{fig:features}
\end{figure*}

The AQ-10, often known as the Autism Spectrum Quotient tool, is a tool for screening designed to determine if an individual should have a comprehensive autism assessment. It consists of 10 questions, and based on their answers, respondents can receive either 0 or 1 point per question. A higher total score indicates a greater likelihood of autism, necessitating further investigation. These datasets are called the AQ-10, a set of 10 adaptive behavioral features for children and adults that take a 10-question screening test. Each question focuses on different domains like communication, attention switching, attention to detail, social interaction, responsiveness, expression, and imagination. The additional attributes of the AQ-10 provide information about the participant taking the autism assessment, including age, gender, ethnicity, jaundice at birth, family members who have Pervasive Developmental Disorder (PDD), country of residence, previous use of the screening app, screening test results, who is taking the test (relation), and class/ASD. The child and adult datasets have the following features that are mentioned in Table \ref{table:feature1_tbl} and Table \ref{table:feature2_tbl}. To identify ASD patients, 10 behavioral traits and 10 different features will be used.

\subsection{Data Pre-processing}
Data pre-processing is the most crucial stage of this study to develop an effective prediction model. It will help us to handle the missing values, remove the duplicate data, and remove each discrepancy in the data before supplying it to the model.

\subsubsection{Missing Value Handling}
The dataset will be loaded into the data frame “df” using the Pandas library. All missing values characterized by '?' will be denoted as “NaN”. The children dataset and the adult dataset have a total of 90 and 192 missing values, respectively. Figure \ref{fig:missing_value} shows where the missing values are in the columns for age, race, and relationships. Missing values can lead to wrong predictions, so the measure of central tendency method (also called the "imputation method") is used to fill in the missing values because the data is less sensitive. The Imputation Method is a fine-tuned and low-computing method to deal with the problem of missing values. We convert the missing values to a co-related non-missing value trend. In this experiment, the missing values were replaced with the median of the non-missing values in the same column using the IMPUTATION (Mean/Median) approach \citep{jadhav2019comparison}.

We used the K-NN imputation method to figure out what to do with the missing values in our datasets. The K-NN functions work on the lines of finding the K nearest neighbor of the missing value or element, then finding and analyzing the ‘feature similarity’ and predicting the missing value. We predict the missing value by looking and finding the k-closest neighbor to it. It is much more accurate and gives more reliable results. A majority of the data is categorical, so we drop the rows that have too many missing values. Because, after trying it out, we found that figuring out all of these missing values and building the model gives us a model that is too well-fitted and less general.

\subsubsection{One-Hot Encoding}
After handling the missing values in the dataset, other preprocessing techniques like One-Hot encoding (for multi-class encoding) are used. We have made the vivid observation that the type of data is non-numeric as well. We have parameters such as country of residence, ethnicity, and relationship to the case. These are all string data types, but they have strong predictive power. We convert these categorical values to non-categorical values via one-hot encoding \citep{potdar2017comparative}. We have used Label Class encoding for converting the Class ASD, which states whether the case has an ASD or not. We have defined to have numeric values of 0 or 1 instead of the string values "Yes" or "No".

\subsubsection{Feature Scaling}
This is a classic normalization technique where the input values are cast in the range of 0-1. It is the coherent range of floating-point values that inhibits the most precision. Normalization for numerical values (using Min Max Scalar) is applied to the dataset to get the desired dataset for the model training. We scale the parameters of age and the result of our dataset. We scale the value of age and result in a range of (0–1). Finally, we end up with highly clean and cohesively preprocessed data \citep{patro2015normalization}. The following equation is typically used to achieve the min-max scaling process:

\begin{equation}
    {p_{scaled}} = \frac{p-min(p)}{max(p)-min(p)}\label{eq-1}
\end{equation}

where $p$ represents input values and $p_{scaled}$ is the normalized value.

\subsection{Feature Selection}
Feature selection techniques are used to reduce the problem of high dimensionality. In this process, only important features are kept, and the less contributing features get removed. Numerous feature selection techniques exist these days. Some of the issues associated with high dimensionality are overfitting, unnecessary noise, high time complexity, etc. Chi-square feature selection is used as a method of feature selection in this study.

\subsubsection{Chi-square Feature Selection}
The Chi-Square selection is used for verifying the independence of events. There are two major occurrences: the existence of a feature and the presence of a class. Its primary function is to determine whether or not a given feature is dependent on a certain class. We take advantage of the fact that the occurrences are not independent if, and only if, that is the case. To ascertain whether or not two variables are interdependent, statisticians use the chi-square test \citep{cai2021application}. The mathematical representation of the Chi-Square formula is given below.

\begin{equation}
    \chi^2 = \sum \frac {(O - E)^2}{E}\label{eq-2}
\end{equation}

Here, the observed count $O$ and anticipated count $E$ may be calculated from data for two independent variables. The Chi-Square test determines the degree of dissimilarity between the two counts, $E$, and $O$. The application of the chi-square technique in feature selection can be deduced readily from its definition. Consider a target variable (class label) and feature variables that characterize each data sample. To determine if there is a relationship between each feature variable and the target variable, we calculate chi-square probabilities.  The feature variable should be discarded when the target variable is independent. When dependent, the feature variable is significant. Figure \ref{fig:features} shows the rank of each feature in terms of Chi-square score for the children, adult, and combined datasets respectively. The figure reveals an intriguing observation about the predominant trait contributing to ASD across various age groups (children and adults). In the case of children, the A4 score, reflecting the challenge of comprehending others' emotions (see Table \ref{table:feature2_tbl}), emerges as the primary trait associated with ASD. Conversely, for adults, the A9 score, indicating a strong inclination to avoid physical contact, emerges as the most influential trait. From all of these features in the corresponding datasets, we select the top 10 features to fit in the selected machine learning algorithms.

\subsection{Train-Test Splitting}
After that, we will divide our datasets into training and testing sets using the idea of holdout sets. The testing set will include unseen data that will be used to evaluate the model's performance, while the training data sets will be employed in fitting the model. By using the $train\_test\_split$ function, we will split the datasets into 70\% train sets and 30\% test sets, respectively. This splitting ratio will give us the best outcome for the corresponding datasets.

\subsection{Cross-validation Technique}
In machine learning, cross-validation is a fundamental model validation approach that determines the model's performance on new and unseen data. To do this, it produces a different dataset known as the cross-validation dataset, which is used to evaluate how well the model performed during training. Cross-validation's main goals are to avoid problems like underfitting and overfitting and to expose model generalization to independent, unseen data.

\subsubsection{k-Fold Cross-validation}
In a standard train-test split, the error measure can vary a lot based on which data points are in the training data set and which are in the test data set. So, the assessment could be different based on how the groups are split up. To solve these challenges, a prominent approach for mode evaluation with the same taste as cross-validation but a little change is called k-fold cross-validation \citep{jung2018multiple}.

\begin{figure*}
\centering
  \includegraphics[width=0.75\linewidth]{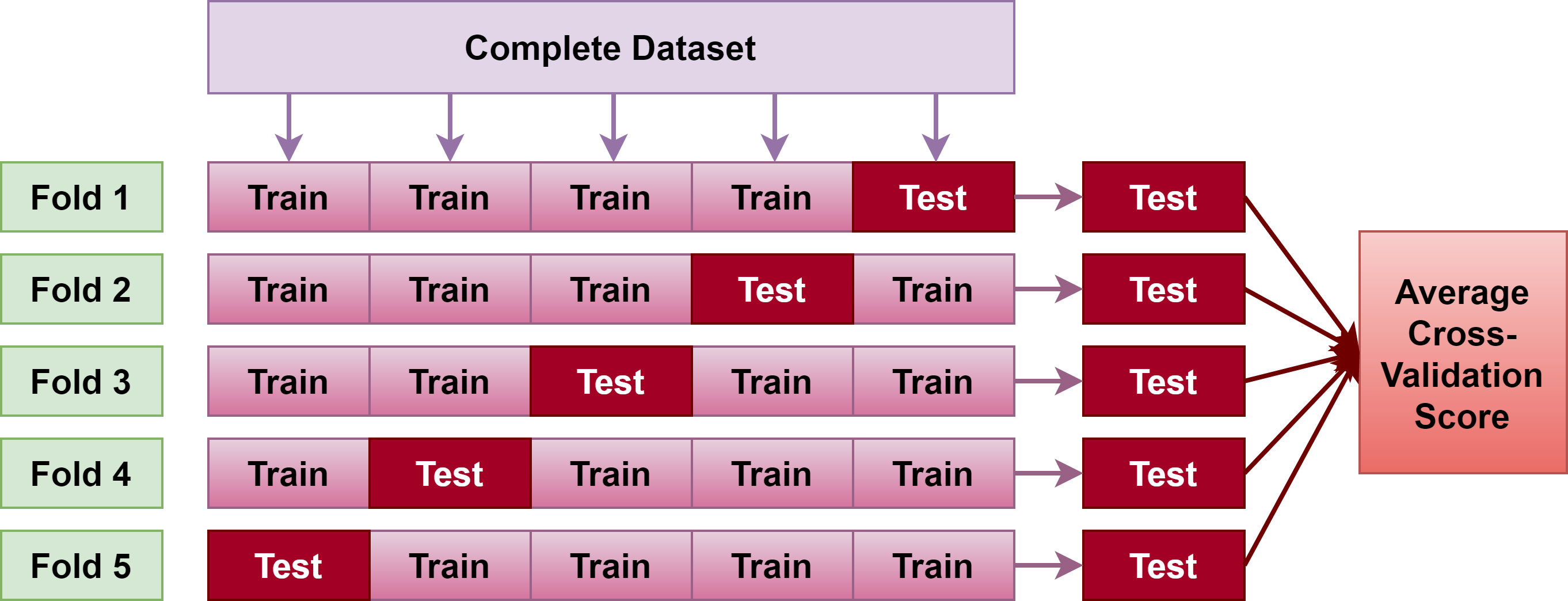}
  \caption{k-fold (k=5) Cross-validation}
  \label{fig:k_fold}
\end{figure*}

K-fold cross-validation is used for model selection and conservative error estimates. The process starts by splitting our data into training and testing sets. Next, k-fold cross-validation divides the training data into k equal sub-buckets (sometimes referred to as groups or folds). Then, $(k-1)$ sub-buckets of training data are utilized to train the learning function, and the remaining bucket (fold) is used for model validation, specifically to figure out an error measure. This procedure is repeated k times for validation with all possible combinations of $(k-1)$ sub-buckets and one bucket. Next, we take the average of the error metrics from all the k different trial runs on the model. The model with the lowest average error score is chosen as the final model. Then, we evaluate the final model against the pre-withheld test data to obtain a conservative error estimation. An illustration is provided to explain the method in Figure \ref{fig:k_fold}.

By using the holdout set method, we are losing a portion of data that could have been used to train the model. As our dataset size is relatively small, we could use the full contribution of the data in the training process. Implementing cross-validation, which involves fitting the model to the data using a series of data subsets, will help us address this issue. Each subset of data is used both for training and validation. A final average validation score is computed after all the data subsets in the sequence are used. Additionally, using this method of validation reduces the risk of overfitting. We will be choosing $k$ as 5 and implementing 5-fold cross-validation.

\subsection{Development of the Classification Models}
Using machine learning methods, it is possible to construct a customized model that can accurately forecast the result of the given dataset. In this research, the model used for the prediction of ASD in children is a supervised classification learning model. We tested the various classifier models on our problem statement and compared the evaluation metrics. Our research makes use of six different prominent models: Naive Bayes, k-Nearest Neighbors (k-NN), Logistic Regression, Support Vector Machine (SVM), Decision Tree, and Random Forest.

\subsubsection{Naive Bayes}
The study of supervised machine learning algorithms is advanced by the application of Naive Bayes (NB), a method that relies on conditional probability (Bayes theorem) and counting.  The assumption is that a feature in a class is independent of any other features in the same class.~\citep{rish2001empirical} One feature affects others. What makes it naïve. Naive Bayes uses joint probability distribution. Naive Bayes extends to multinomial. It predicts by calculating the mean, standard deviation, and class probabilities of the training data, making it easy to use. We used the Multinomial Naive Bayes model for our research among the other types of Naive Bayes models in our binary classification problem. In this model, we assume that the attributes are conditionally independent and figure out the class conditional probability.

\subsubsection{k-Nearest Neighbors}
This algorithm, also known as the "lazy learner algorithm,")~\citep{zhang2005k},~\citep{peterson2009k} attempts to identify a data point by looking at the neighboring data points around it without generating a model of the data beforehand. In order to determine the nearest neighbors, the distance between the existing data points and the new data point from the data is calculated. It then polls the data point's neighbors and makes a decision for classification based on a majority vote. For predicting the class, it analyses the closest k instances and the value predicted will be the most common class. We label a group of points and use them to label other points. After labeling a new point, it looks at the other points and finds out the k closest points. Each point votes, so any point is labeled according to the vote. So, to figure out the class of a new instance, we need to search for its closest instances. The closest neighbor is determined using the Euclidean distance formula. The formula for this is just the root square of the difference in squares between the new and existing instances.

\subsubsection{Logistic Regression}
Linear models are statistical models that aim to establish a relationship between a dependent response variable and one or more independent factors. A popular linear statistical model in discriminant analysis is logistic regression.~\citep{hosmer2013applied} It is a statistical analysis approach that predicts a data point based on previous observations. It determines whether an independent variable has an effect on a binary dependent variable, implying that there are only two potential outcomes given some input. It provides accuracy, making the model easier to understand. Logical regression works best when there is little or no multicollinearity between the independent factors. We selected this model because the equation is easy to figure out and the factors tell us how the independent variable is affected.

\subsubsection{Support Vector Machine (SVM)}
Support Vector Machine (SVM) classifies data into two groups. In a high-dimensional space, they represent distinct classes and locate a hyperplane with the maximum margin between their data points. SVM training creates a boundary between the two categories, and fresh data points are identified by their position \citep{zhang2012support}. SVM is a straightforward method for identifying Autism Spectrum Disorder.

\subsubsection{Decision Tree}
It is a tool for making decisions that employs a tree-structure flowchart-like appearance. It mainly focuses on all the decisions, their outputs, and possible results. This will fall under both regression and classification. It is used mostly in classification and sometimes in regression as well. To predict whether the person had autistic traits or not, we chose a decision tree classifier for the prediction model. The whole dataset is passed to the tree root node, and then the data is split using features. This process is going to continue recursively until the node gets a unique label. We can split the algorithm into two phases, in which the first phase is building the tree and the second is classifying the test data. We will first import the Decission Tree (DT) classifier \citep{song2015decision} from the scikit-learn (sklearn) library into Python, and then we will select the best features for constructing the tree. After this, we iterate over each data feature by visualizing the graph and checking its maximum information gain (IG). If its value is 0, that means it is pure and will return a leaf node. If not, then it will be split into two parts (false and true). The function is going to run recursively, and the decision tree will be formed from the branches. After the construction of the tree, we classify the test data according to the leaf node prediction. We iterate over the tree, and it reaches the leaf node and gets classified.

\subsubsection{Random Forest}
A tree-based approach called Random Forest improves prediction accuracy by assembling uncorrelated trees using methods like bagging and feature randomization ~\citep{breiman2001random}.  By averaging predictions (for regression) or votes (for classification) from several trees at terminal leaf nodes, bootstrap aggregation is used to reduce variance. The accuracy of a single decision tree is increased by generating numerous trees, each selected randomly from the training set without replacement.  However, this increased accuracy comes at the expense of decreased interpretability. We will import the random forest classifier from the scikit-learn ensemble library into Python and fit the training data into it.

\subsubsection{Extreme Gradient Boosting}
Extreme Gradient Boosting, or XGBoost, is a robust machine learning method that leverages decision trees and gradient boosting. XGBoost eliminates the need for manual data preparation because of its ability to automatically deal with missing data during training and prediction. It allows for cross-validation methods to be used to predict the model's behavior on unobserved data and improve the model's hyperparameters. In data science, XGBoost is extensively utilized because it efficiently resolves massive problems with limited resources \citep{chen2016xgboost}. To predict the results of dataset $S$ containing $p$ samples and $q$ attributes, the XGBoost model uses $m$ additive functions. Where $x_{i}$ is a vector in ${R}^{q}$ and $y_{i}$ is a real integer, this is the representation of the dataset.
   $ S = \left \{ (x_{i}, y_{i}) \right \} $ where, $ \left | S \right |= p, x_{i}\epsilon \mathbb{R}^{q}, y_{i}\epsilon \mathbb{R}^{q} $.
   
\begin{equation}
\begin{aligned}
\label{eq:four}
\hat{y}_{i} = \O (x_{i}) = \sum_{m=1}^{M} f_{m}(x_{i}), f_{m}\epsilon \left \{ f(x)=w_{n}(x)) \right \}\\ 
(n: \mathbb{R}^{q}\rightarrow T, w\rightarrow \mathbb{R}^{T})
\end{aligned}
\end{equation}

where $n$ represents the structure of each tree that maps a guide to the corresponding leaf nodes, and $T$ denotes the number of leaves. For every function $f_{m}$, there is a distinct structure of trees $n$ and leaf weights $w$.

\subsubsection{Artificial Neural Network (ANN)}
A mathematical model that attempts to depict how the mind functions is called a neural network. A grouping of activation functions and perceptrons is called an artificial neural network (ANN). By connecting the perceptrons, hidden layers, or units, are created. Artificial neural networks, as they are commonly known, are created by mapping the input layers to the output layers in a lower-dimensional space using hidden units on a non-linear basis. An ANN) is a type of input-to-output map. The weighted sum of the biased inputs yields the map. Models refer to the weight and bias values as well as the architecture. The ANN has several parameters that may be optimized. The computed error is used to update the weights retroactively. Optimization is the process of minimizing errors \citep{mishra2014view}. In this study, the proposed ANN model consists of 4 dense layers, where the dense layers contain 1024, 512, and 512 neurons, respectively, and ReLu is used as the activation function. There are also three batch normalization layers and an output layer.

\subsection{Development of the Clustering Models}
\subsubsection{K-Means Algorithms}
K-means clustering iterates until it finds the best centroid. The number of clusters is assumed to be known at this time. The 'K' in K-means represents the algorithm's overall cluster count. Also known as the flat clustering algorithm.
The objective function is:
\begin{equation}
\begin{aligned}
\label{eq:four}
\hat{F} = \sum_{i=1}^{m} \sum_{j=1}^{K} w_{ij} ||x^{i}-q_{j}||^{2}, 
\end{aligned}
\end{equation}
The value of $w_{ij}$ is equal to 1 if $x^i$ belongs to cluster $j$, otherwise, it is 0. Here, $q_j$ represents the cluster centroid of $x^i$.For the K-Means clustering method, the runtime complexity is denoted as $O(n^2d)$. Here, $n$ represents the overall number of data points being clustered, while $d$ corresponds to the number of dimensions.  It takes time proportionate to the square of the number of data points for the technique to compute the distance between each pair of data points \citep{ikotun2023k}.

\subsubsection{Agglomerative}
Agglomerative clustering, a well-known data mining technique, is a bottom-up approach that works by repeatedly combining groups that have similarities; the resultant hierarchy may be shown in a dendrogram. In this method, each data point is treated as its own cluster. As the algorithm progresses up the hierarchy, it merges pairs of clusters. 
In the most fundamental form of agglomerative clustering, the algorithm must compute the distance matrix for each pair of data points, which requires $O(n^2)$ time. It then brute-forces the nearest pair of clusters based on the linking requirement in each iteration. As a result, the runtime complexity of agglomerative clustering is $O(n3)$, where $n$ represents the total number of data points \citep{mullner2011modern}.

\subsubsection{GMM}
A Gaussian Mixture Model (GMM) is a probabilistic model that uses soft grouping to group data into various groups.
 It is especially useful when the boundaries between clusters are not clear. GMM figures out how likely it is that a data point belongs to each cluster by assuming that each cluster has a Gaussian distribution with a mean vector $(\mu)$ and a correlation matrix $(\Sigma)$. It has two main parts: the means and the covariances. GMM uses the Expectation-Maximization (EM) method to group unidentified data in the same way that k-means does. EM for GMM takes $O(nkdT)$ runtime complexity, where $n$ represents the total amount of data points, $k$ denotes the number of clusters, $d$ signifies the number of dimensions, and $T$ corresponds to the total number of iterations required for convergence \citep{patel2020clustering}.

\subsubsection{Spectral Clustering}
The concept of spectral clustering may be traced back to graph theory, where it was first utilized to find groups of connected nodes. In addition to graph data, we can also use this approach to cluster other types of data.
The eigenvalues (spectrum) of unique matrices constructed from the graph or data set are used in spectral clustering.
When a cluster's center and spread cannot adequately describe the whole cluster, for as when clusters are nested circles on a 2D plane, or more broadly when a cluster's structure is significantly non-convex, spectral clustering is particularly beneficial. 
Where the number of clusters is $k$ and the number of dimensions in the data is $d$ , the complexity of the spectral clustering, is denoted as $O(n^3 + kn2d)$ \citep{von2007tutorial}. 

\subsubsection{BIRCH}
The BIRCH clustering technique was developed for use in machine learning and is optimized for large-scale data sets. This approach avoids scanning every point in a dataset to conduct clustering.
BIRCH clusters the dataset into tiny summaries before clustering them. It clusters data indirectly. BIRCH is widely used alongside other clustering methods because after summarizing, the summary may be clustered. It creates a tree data structure and reads leaf centroids. These centroids may be used for Agglomerative Clustering or as the cluster centroid \citep{zhang1997birch}.
The BIRCH clustering algorithm is often quicker than other clustering algorithms because of its quadratic time complexity, with a runtime complexity of $O(n log n + nb log(T) + nT)$.

\begin{sidewaystable*}
\caption{Optimized hyperparameters of eight ML models for the corresponding dataset.
\vspace{1.0 in}}
\renewcommand{\arraystretch}{1.7}
\label{tab:parameters}
\begin{center}
{\footnotesize
\begin{tabular}{ccccc}
\hline
\multirow{2}{*}{\textbf{Models Name}} & \multirow{2}{*}{\textbf{Initial Parameters}}                                                                                                                                                                                                                                                                                                                                                   & \multicolumn{3}{c}{\textbf{Optimized Parameters}}                                                                                                                                                                                                                                                                                                                                                                                                                                                                                                                                                                                            \\ \cline{3-5} 
                             &                                                                                                                                                                                                                                                                                                                                                                                       & \multicolumn{1}{c}{\textbf{Children dataset}}                                                                                                                                                                             & \multicolumn{1}{c}{\textbf{Adult dataset}}                                                                                                                                                                                & \textbf{Combined dataset}                                                                                                                                                                             \\ \hline
\textbf{NB}                           & 'var\_smoothing': np.logspace(0,-9, num=100)                                                                                                                                                                                                                                                                                                                                          & \multicolumn{1}{c}{'var\_smoothing': 0.4328}                                                                                                                                                                     & \multicolumn{1}{c}{'var\_smoothing': 0.2848}                                                                                                                                                                     & 'var\_smoothing': 0.0284                                                                                                                                                                     \\ \hline
\textbf{KNN}                          & \begin{tabular}[c]{@{}c@{}}'n\_neighbors': {[}1, 112, 223, 334, 445, 556, 667, 778, 889, 1000{]}, \\ 'weights': {[}'uniform', 'distance'{]}, \\ 'algorithm': {[}'auto', 'ball\_tree', 'kd\_tree', 'brute'{]}, \\ 'leaf\_size': {[}1, 112, 223, 334, 445, 556, 667, 778, 889, 1000{]}\end{tabular}                                                                                     & \multicolumn{1}{c}{\begin{tabular}[c]{@{}c@{}}weights='uniform', \\ n\_neighbors= 1, \\ leaf\_size=445, \\ algorithm='brute'\end{tabular}}                                                                       & \multicolumn{1}{c}{\begin{tabular}[c]{@{}c@{}}weights='distance', \\ n\_neighbors= 1, \\ leaf\_size=112, \\ algorithm='ball\_tree'\end{tabular}}                                                                 & \begin{tabular}[c]{@{}c@{}}weights='distance', \\ n\_neighbors= 1, \\ leaf\_size=112, \\ algorithm='ball\_tree'\end{tabular}                                                                 \\ \hline
\textbf{SVM}                          & \begin{tabular}[c]{@{}c@{}}'C': {[}1, 3, 5, 7, 9, 11, 13, 15, 17, 20{]}, 'kernel': {[}'linear', 'poly', 'rbf', 'sigmoid'{]}, \\ 'degree': {[}1, 2, 3, 4, 5, 6, 7, 8, 9, 10{]}\end{tabular}                                                                                                                                                                                            & \multicolumn{1}{c}{\begin{tabular}[c]{@{}c@{}}kernel='linear',\\ degree=3, C=13\end{tabular}}                                                                                                                    & \multicolumn{1}{c}{\begin{tabular}[c]{@{}c@{}}kernel='linear',\\ degree=3, C=13\end{tabular}}                                                                                                                    & \begin{tabular}[c]{@{}c@{}}kernel='linear',\\ degree=3, C=13\end{tabular}                                                                                                                    \\ \hline
\textbf{RF}                           & \begin{tabular}[c]{@{}c@{}}'n\_estimators': {[}200, 400, 600, 800, 1000, 1200, 1400, 1600, 1800, 2000{]},\\  'max\_features': {[}'auto', 'sqrt', 'log2'{]}, \\ 'max\_depth': {[}10, 120, 230, 340, 450, 560, 670, 780, 890, 1000{]},\\  'min\_samples\_split': {[}2, 5, 10, 14{]},\\  'min\_samples\_leaf': {[}1, 2, 4, 6, 8{]}, \\ 'criterion': {[}'entropy', 'gini'{]}\end{tabular} & \multicolumn{1}{c}{\begin{tabular}[c]{@{}c@{}}n\_estimators = 800, \\ min\_samples\_split = 2, \\ min\_samples\_leaf = 4, \\ max\_features = 'log2', \\ max\_depth = 450, \\ criterion = 'entropy'\end{tabular}} & \multicolumn{1}{c}{\begin{tabular}[c]{@{}c@{}}n\_estimators = 800, \\ min\_samples\_split = 2, \\ min\_samples\_leaf = 4, \\ max\_features = 'log2', \\ max\_depth = 450, \\ criterion = 'entropy'\end{tabular}} & \begin{tabular}[c]{@{}c@{}}n\_estimators = 800, \\ min\_samples\_split = 5, \\ min\_samples\_leaf = 1, \\ max\_features = 'sqrt', \\ max\_depth = 560, \\ criterion = 'entropy'\end{tabular} \\ \hline
\textbf{DT}                           & \begin{tabular}[c]{@{}c@{}}"criterion":{[}'gini','entropy'{]}, \\ "max\_depth":(150, 155, 160), \\ "min\_samples\_split":range(1,10),\\  "min\_samples\_leaf":range(1,5)\end{tabular}                                                                                                                                                                                                 & \multicolumn{1}{c}{\begin{tabular}[c]{@{}c@{}}min\_samples\_split = 2, \\ min\_samples\_leaf = 1, \\ max\_depth = 155, \\ criterion = 'gini'\end{tabular}}                                                       & \multicolumn{1}{c}{\begin{tabular}[c]{@{}c@{}}min\_samples\_split = 3,\\  min\_samples\_leaf = 1, \\ max\_depth = 150, \\ criterion = 'entropy'\end{tabular}}                                                    & \begin{tabular}[c]{@{}c@{}}min\_samples\_split = 3, \\ min\_samples\_leaf = 3, \\ max\_depth = 160, \\ criterion = 'gini'\end{tabular}                                                       \\ \hline
\textbf{XGB}                          & \begin{tabular}[c]{@{}c@{}}'max\_depth': range (2, 10, 1), \\ 'n\_estimators': range(60, 220, 40), \\ 'learning\_rate': {[}0.1, 0.01, 0.05{]}\end{tabular}                                                                                                                                                                                                                            & \multicolumn{1}{c}{\begin{tabular}[c]{@{}c@{}}n\_estimators = 140, \\ max\_depth = 4, \\ learning\_rate = 0.1\end{tabular}}                                                                                      & \multicolumn{1}{c}{\begin{tabular}[c]{@{}c@{}}n\_estimators = 140, \\ max\_depth = 4, \\ learning\_rate = 0.1\end{tabular}}                                                                                      & \begin{tabular}[c]{@{}c@{}}n\_estimators = 140, \\ max\_depth = 2, \\ learning\_rate = 0.1\end{tabular}                                                                                      \\ \hline
\textbf{LR}                           & 'penalty': {[}'l1','l2'{]}, 'C': {[}0.001,0.01,0.1,1,10,100,1000{]}                                                                                                                                                                                                                                                                                                                   & \multicolumn{1}{c}{penalty = 'l2', C = 1000}                                                                                                                                                                     & \multicolumn{1}{c}{penalty = 'l2', C = 100}                                                                                                                                                                      & penalty = 'l2', C = 100                                                                                                                                                                      \\ \hline
\textbf{ANN}                          & \begin{tabular}[c]{@{}c@{}}optimizer='SGD, adam, RMSprop', \\ loss='binary\_crossentropy', \\ batch\_size=8, 16, 32, 64, \\ factor=0.8, patience=10\end{tabular}                                                                                                                                                                                                                      & \multicolumn{3}{c}{\begin{tabular}[c]{@{}c@{}}optimizer='adam', loss='binary\_crossentropy', \\ batch\_size=32, factor=0.8, patience=10\end{tabular}}                                                                                                                                                                                                                                                                                                                                                                                                                                                                               \\ \hline
\end{tabular}
}
\end{center}
\end{sidewaystable*}

\section{Evaluation Metrics}
\subsection{Classification metrics}
Various classification metrics such as precision, recall, sensitivity, F1-score, accuracy, Cohen's Kappa Score, AUC, and log loss, are used to measure the final prediction of the model \citep{ahamad2022early}. These metrics assess the model's performance on a testing set, facilitating comparisons and drawing conclusions on the model's predictive capability.

\begin{itemize}
    \item{\textbf{Precision:}} Precision is the ratio of positive occurrences through total actual positive instances. Simply, precision indicates how accurate a model is when it claims to be correct. To calculate the same thing, the following formula is used:

    \begin{equation}
        {Precision} = \frac{TP}{TP+FP} \label{eq}
    \end{equation}

    \item{\textbf{Recall or Sensitivity:}} Recall is a measure that quantifies the proportion of actual positive instances correctly identified by a system or model, expressed as a ratio of true positives to the total number of real positive instances. The following formula is used to calculate the result.

    \begin{equation}
        {Recall} = \frac{TP}{TP+FN} \label{eq}
    \end{equation}

    \item{\textbf{Specificity:}} Specificity measures how effectively a model recognizes negative occurrences among all real negative occurrences in a test. It is the number of true negatives compared to the total number of true negatives that have been analyzed. Here is the formula to calculate specificity:
    
    \begin{equation}
        {Specificity} = \frac{TN}{TN+FP} \label{eq}
    \end{equation}

    \item{\textbf{Accuracy:}} In machine learning, accuracy is a measure of how many data points are predicted correctly out of the total. It represents the percentage of correct predictions in a test dataset. It is essential because AI learns autonomously, making it difficult to determine whether or not input data is accurate.

    \begin{equation}
        {Accuracy} = \frac{TP+TN}{TP+TN+FP+FN} \label{eq}
    \end{equation}

    \item{\textbf{F1\- Score:}} F1-score is a combined measure of precision and recall, where 1 is the greatest possible result and 0 is the lowest. A high F1 score suggests accurate threat detection with a balance between false positives and false negatives. The F-beta score allows us to discover the optimal compromise between precision and recall in certain circumstances.

    \begin{equation}
        {F1 \; score} = \frac{2*Recall*Precision}{Recall+Precision} \label{eq}
    \end{equation}

    \item{\textbf{Cohen’s Kappa Score:}} The Kappa score indicates the interrelationship between the two categorical classes considered for classification, which in this instance are mask and no mask. It provides insight into the robustness of the model since it provides the possibility of classification that occurred by chance. This value is obtained by formulating the values with the formula given below.

    \begin{equation}
        {Kappa \; Score} = \frac{p_0-p_e}{1-p_e} \label{eq}
    \end{equation}

    where $P_0$ is the value of the observed outcome, which has the same value as accuracy, and $P_e$ is the probability of hypothetically getting the desired outcome.
 
    \item{\textbf{ROC-AUC:}} The Receiver Operating Characteristic (ROC) curve illustrates binary predictor performance at various decision levels. It shows how true positives and false positives balance each other out. The area under the curve, often known as the AUC, is a method to summarize the curve. Higher AUC numbers mean that the model is doing better. AUC values between 0.7 and 0.8 are good, 0.8 to 0.9 are great, and anything above 0.9 is amazing. Compared to accuracy, AUC is a better measure of a classifier's performance.
    \item{\textbf{Log Loss:}} Log loss is a fundamental measure for probability-estimated classification models. Calculating the negative average of the logarithm of the corrected predicted probability for each case assesses accuracy. While raw log-loss data are difficult to read, they are useful for model comparison. In practice, lower log loss means better problem-solving prediction.

     \begin{equation}
     \begin{aligned}
         \label{eq:logloss}
         H_{p}(q) = -\frac{1}{n}\sum_{i=1}^{n}x_{i}.log(p(x_{i})) \\+(1-x_{i}).log(1-p(x_{i}))
     \end{aligned}
    \end{equation}

Where $x$ represents the target variable level, $p(x)$ is the projected point probability for the target value, and $H(q)$ is the estimated value of log loss.
 
\end{itemize}

\subsection{Clustering metrics}

To assess how well different clustering algorithms perform, we use an unsupervised metric called the Silhouette Coefficient (SC)~\citep{rousseeuw1987silhouettes}. This metric scores clustering quality on a scale from -1 to 1. A score of 1 means the clustering is perfect, -1 means it's completely incorrect, and a value near 0 suggests some overlapping clusters. We also employ two other metrics: Normalized Mutual Information (NMI)~\citep{vinh2009information} \& Adjusted Rand Index (ARI)~\citep{hubert1985comparing}. NMI measures how well the obtained clusters align with the true classes, with 1 being a perfect match and 0 indicating entirely wrong labels. ARI also ranges from -1 to 1, and its interpretation is similar to SC.

\subsection{Implementation Details}
The entire experiment was performed on Jupyter Notebook in Google Colaboratory which is Google's cloud-based service.

\section{Results and discussion}
The model assessment findings for the chosen datasets are presented and discussed in this section. The training and measurement of the models using the classification report and confusion matrix determine the assessment outcomes in terms of precision, recall or sensitivity, specificity, accuracy, f1-score, Cohen Kappa score, ROC-AUC, and log loss.

After completing the data preprocessing, we have to find out the best hyperparameters of the machine learning model by using the hyperparameter optimization approach. Because the best parameters of the models will show the best performance of the models. After finding out the best parameters of the models, we will perform the model training stage by utilizing the best model parameters. After completing the training stage, we will save the trained model, and after that, the saved model will be applied to the test sets and the performance of the ML models will be calculated. In the following Table \ref{tab:parameters}, the best hyperparameters of the models are shown.

\subsection{Classification results}
After applying k-fold cross-validation techniques to the datasets, it gives a better indication of accuracy with less overfitting in the selected machine learning algorithms. In this experiment, eight classification algorithms have been selected for the prediction of the final results. After using eight separate models and using specific evaluation metrics to draw conclusions and make comparisons, we look at the results of the models how they tend to deal with different data, and how they give different optimized results and certain changed conditions. Table \ref{tab:child} shows the accuracy, precision, recall, specificity, F1-score, AUC Score, Kappa score, and log loss score of six different machine learning models for the ASD Detection Children dataset. Table \ref{tab:adult} and Table \ref{tab:combined} show the same metrics for ASD Detection Adult and ASD Detection Combined datasets respectively. 

\begin{table*}[htbp]
\caption{Accuracy, Precision, Recall, Specificity, F1-score, AUC Score, Kappa score \& Log loss score of six different machine learning models for ASD Detection Children dataset. ({\em bold values are better}).
\vspace{-0.15in}}
\label{tab:child}
\begin{center}
\setlength{\tabcolsep}{4pt}
\renewcommand{\arraystretch}{1}
{\footnotesize
\begin{tabular}{ccccccccc}
\toprule
\multicolumn{1}{l}{\bf Model}  &\multicolumn{1}{c}{\bf Accuracy (\%)} &\multicolumn{1}{c}{\bf Precision (\%)} &\multicolumn{1}{c}{\bf Recall (\%)} &\multicolumn{1}{c}{\bf Specificity (\%)} &\multicolumn{1}{c}{\bf F1-score (\%)} & \multicolumn{1}{c}{\bf AUC (\%)} & \multicolumn{1}{c}{\bf Kappa (\%)} &\multicolumn{1}{c}{\bf Log Loss}\\
\midrule
Naive Bayes & 94.25& 90.91 & 97.56& 91.30& 94.12& 94.43& 88.51& 2.071 \\
k-Nearest Neighbors & 86.21& 87.18& 82.93& 89.14& 85.00& 86.03& 72.25& 4.972 \\
Support Vector Machine & \textbf{100}& \textbf{100}&\textbf{100}& \textbf{100}& \textbf{100}& \textbf{100}& \textbf{100}& \textbf{0.00} \\
Random Forest  & 96.55& 93.18& \textbf{100}& 93.48& 96.47& 96.74& 93.11& 1.243 \\
Decision Tree & 87.36& 87.50& 85.37& 89.13& 86.42& 87.25& 74.60& 4.557 \\
Extreme Gradient Boosting & 97.70& 97.56& 97.56& 97.83& 97.56& 97.69& 91.60& 0.829 \\
Logistic Regression  & \textbf{100}& \textbf{100}& \textbf{100}& \textbf{100}& \textbf{100}& \textbf{100}& \textbf{100}& \textbf{0.00}\\
Artificial Neural Network & 98.85& 97.62& \textbf{100}& 97.83& 98.80& 98.91& 97.70& 0.414 \\
\bottomrule
\end{tabular}
}
\end{center}
\end{table*}
\begin{table*}[htbp]
\caption{Accuracy, Precision, Recall, Specificity, F1-score, AUC Score, Kappa score \& Log loss score of six different machine learning models for ASD Detection Adult dataset. ({\em bold values are better}).
\vspace{-0.15in}}
\label{tab:adult}
\begin{center}
\setlength{\tabcolsep}{4pt}
\renewcommand{\arraystretch}{1}
{\footnotesize
\begin{tabular}{ccccccccc}
\toprule
\multicolumn{1}{l}{\bf Model}  &\multicolumn{1}{c}{\bf Accuracy (\%)} &\multicolumn{1}{c}{\bf Precision (\%)} &\multicolumn{1}{c}{\bf Recall (\%)} &\multicolumn{1}{c}{\bf Specificity (\%)} &\multicolumn{1}{c}{\bf F1-score (\%)} & \multicolumn{1}{c}{\bf AUC (\%)} & \multicolumn{1}{c}{\bf Kappa (\%)} &\multicolumn{1}{c}{\bf Log Loss}\\
\midrule
Naive Bayes & 96.19& 95.00& 91.94& 97.98& 93.44& 94.95& 90.76& 1.373\\
k-Nearest Neighbors & 89.05& 84.21& 77.42& 93.92& 80.67& 85.67& 73.05& 3.948\\
Support Vector Machine & 96.19& 93.55& 93.55& 97.30& 93.55& 95.42& 90.85& 1.373\\
Random Forest & 95.71& 94.92& 90.32& 97.97& 92.56& 94.15& 89.55& 1.545 \\
Decision Tree & 87.14& 84.31& 69.35& 94.59& 76.11& 81.97& 80.02& 4.634 \\
Extreme Gradient Boosting & 96.19& 95.00& 91.94& 97.97& 93.44& 94.95& 90.76& 1.373\\
Logistic Regression & \textbf{97.14}& \textbf{96.67}& 93.55& \textbf{98.65}& \textbf{95.08}& 96.09& \textbf{93.07}& \textbf{1.029}\\
Artificial Neural Network & 96.67& 91.04& \textbf{98.39}& 95.95& 94.57& \textbf{97.17}& 92.17& 1.201\\
\bottomrule
\end{tabular}
}
\end{center}
\end{table*}

\begin{table*}[htbp]
\caption{Accuracy, Precision, Recall, Specificity, F1-score, AUC Score, Kappa score \& Log loss score of six different machine learning models for ASD Detection Combined dataset. ({\em bold values are better}).}
\label{tab:combined}
\begin{center}
\setlength{\tabcolsep}{4pt}
\renewcommand{\arraystretch}{1}
{\footnotesize
\begin{tabular}{ccccccccc}
\toprule
\multicolumn{1}{l}{\bf Model}  &\multicolumn{1}{c}{\bf Accuracy (\%)} &\multicolumn{1}{c}{\bf Precision (\%)} &\multicolumn{1}{c}{\bf Recall (\%)} &\multicolumn{1}{c}{\bf Specificity (\%)} &\multicolumn{1}{c}{\bf F1-score (\%)} & \multicolumn{1}{c}{\bf AUC (\%)} & \multicolumn{1}{c}{\bf Kappa (\%)} &\multicolumn{1}{c}{\bf Log Loss}\\
\midrule
Naive Bayes & 93.27& 86.41 & \textbf{93.68}& 93.07& 89.89& 93.38& 84.86& 2.427 \\
k-Nearest Neighbors & 83.16& 72.73& 75.79& 86.63& 74.23& 81.21& 61.73& 6.068 \\
Support Vector Machine & 91.58& 88.04& 85.26& 94.55& 86.63& 89.91& 80.49& 3.033\\
Random Forest & 93.27& \textbf{92.13}& 86.32& \textbf{96.53}& 89.13& 91.43& 84.26& 2.427 \\
Decision Tree & 91.92& 87.37& 87.37& 94.06& 87.37& 90.71& 81.43& 2.913 \\
Extreme Gradient Boosting & 92.59& 91.01& 85.26& 96.04& 88.04& 90.65& 82.69& 2.669 \\
Logistic Regression & 91.58& 91.67& 81.05& \textbf{96.53}& 86.03& 88.79& 80.04& 3.034 \\
Artificial Neural Network & \textbf{94.28}& 89.80& 92.63& 95.05& \textbf{91.19}& \textbf{93.84}& \textbf{86.95}& \textbf{2.063}\\
\bottomrule
\end{tabular}
}
\end{center}
\end{table*}

\begin{figure*}[]
\centering

    \includegraphics[width=.49\textwidth]{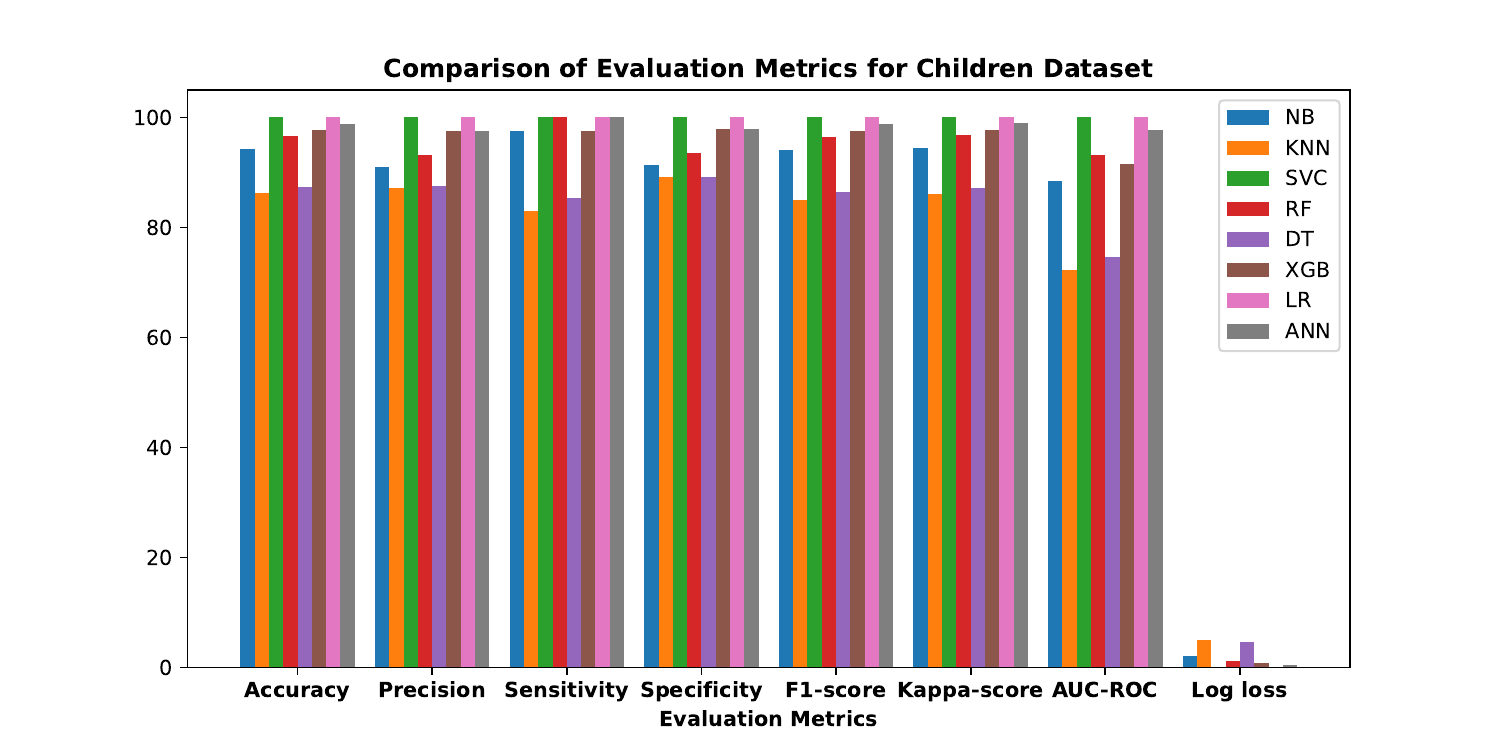}
    \includegraphics[width=.49\textwidth]{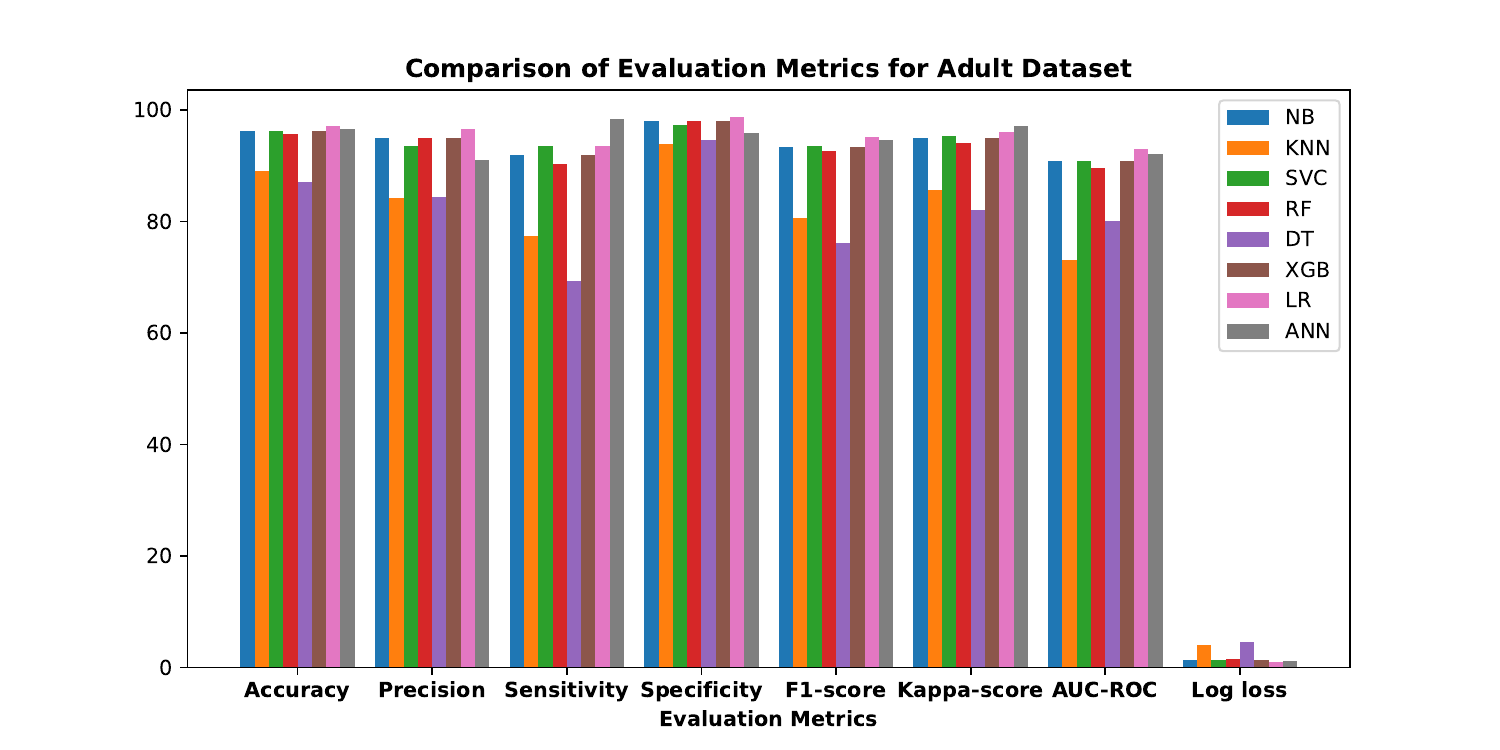}
    \includegraphics[width=.49\textwidth]{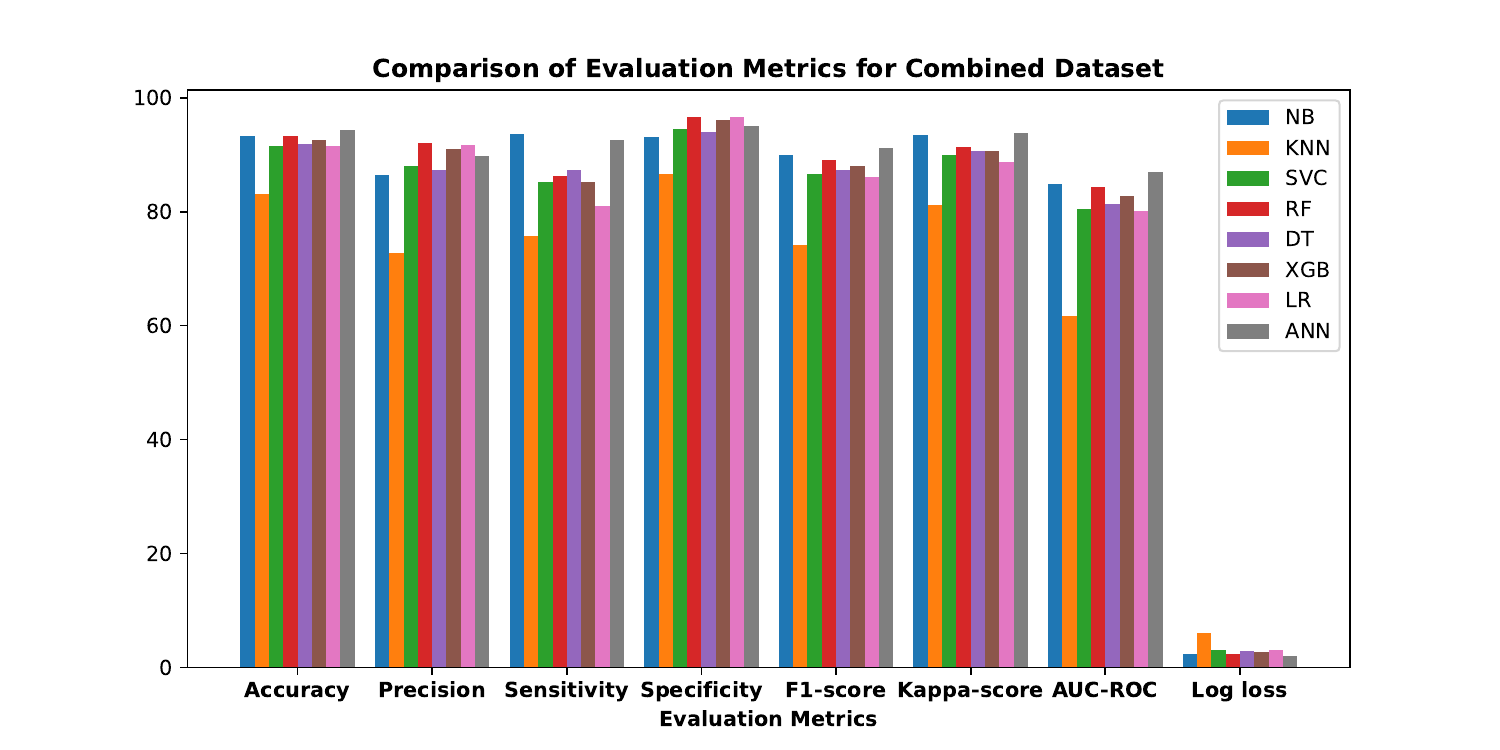}

\caption{Samples of the Image Dataset}
\label{fig-1}
\end{figure*}

\begin{figure*}
\centering
    \includegraphics[width=.24\textwidth]{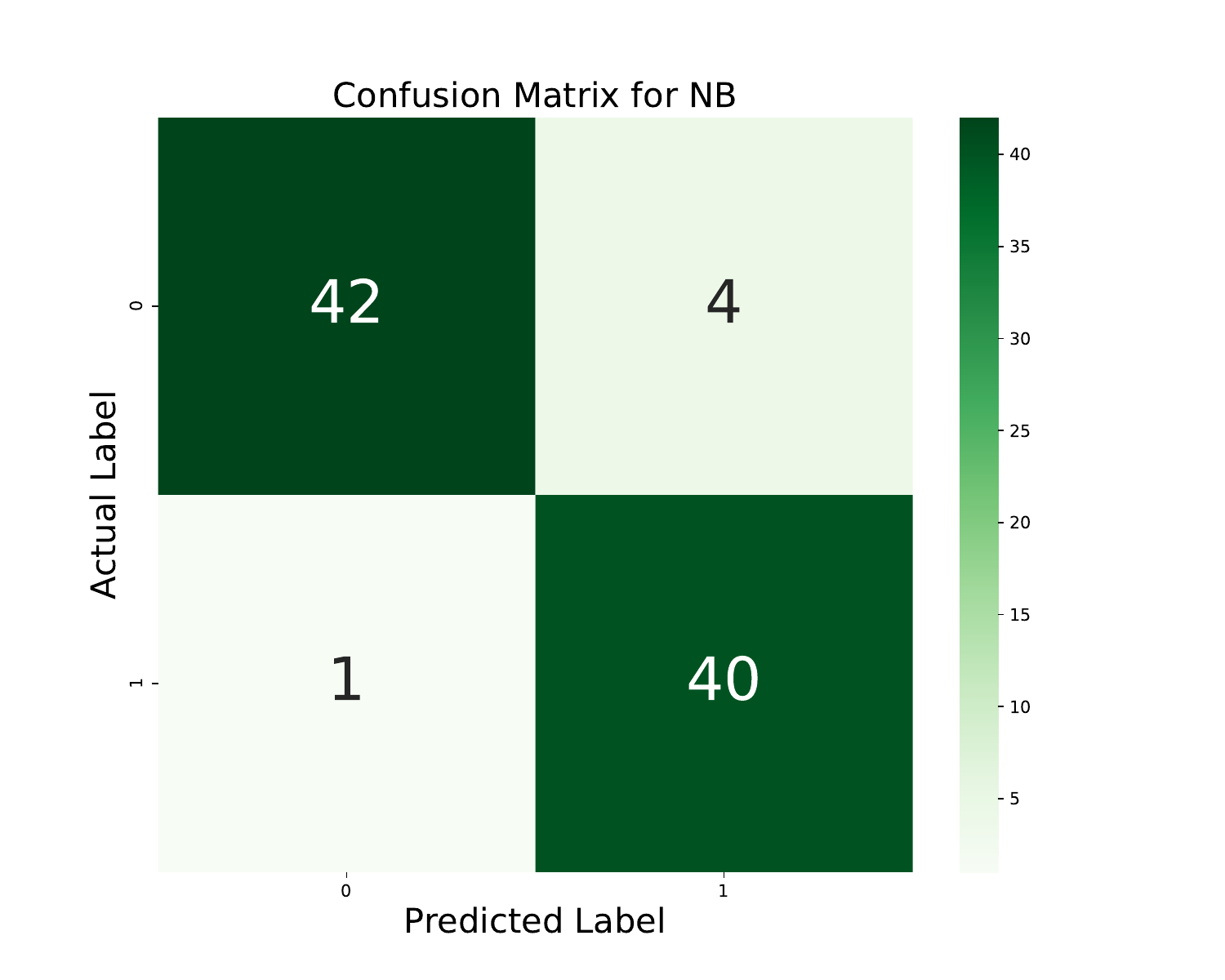}
    \includegraphics[width=.24\textwidth]{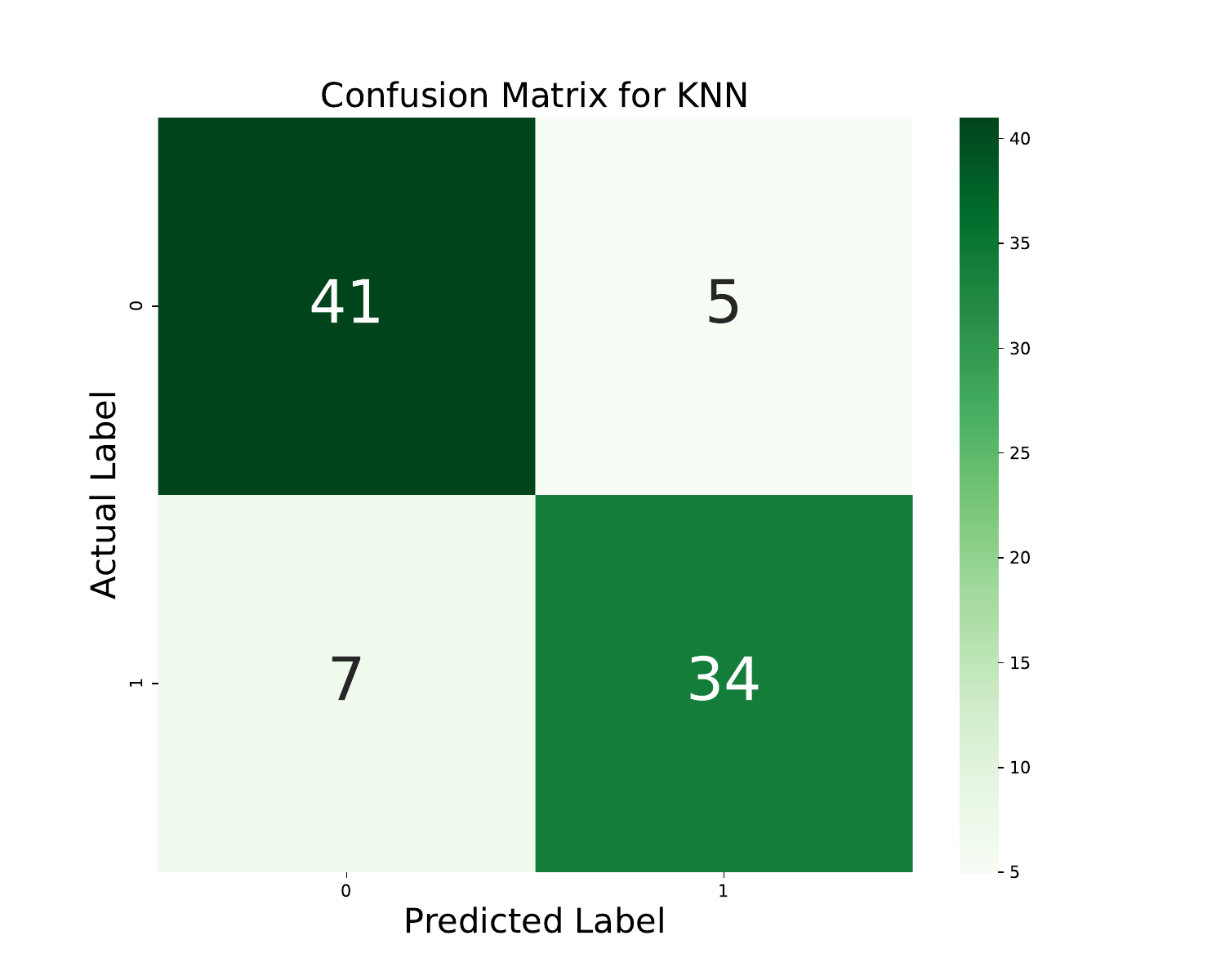}
    \includegraphics[width=.24\textwidth]{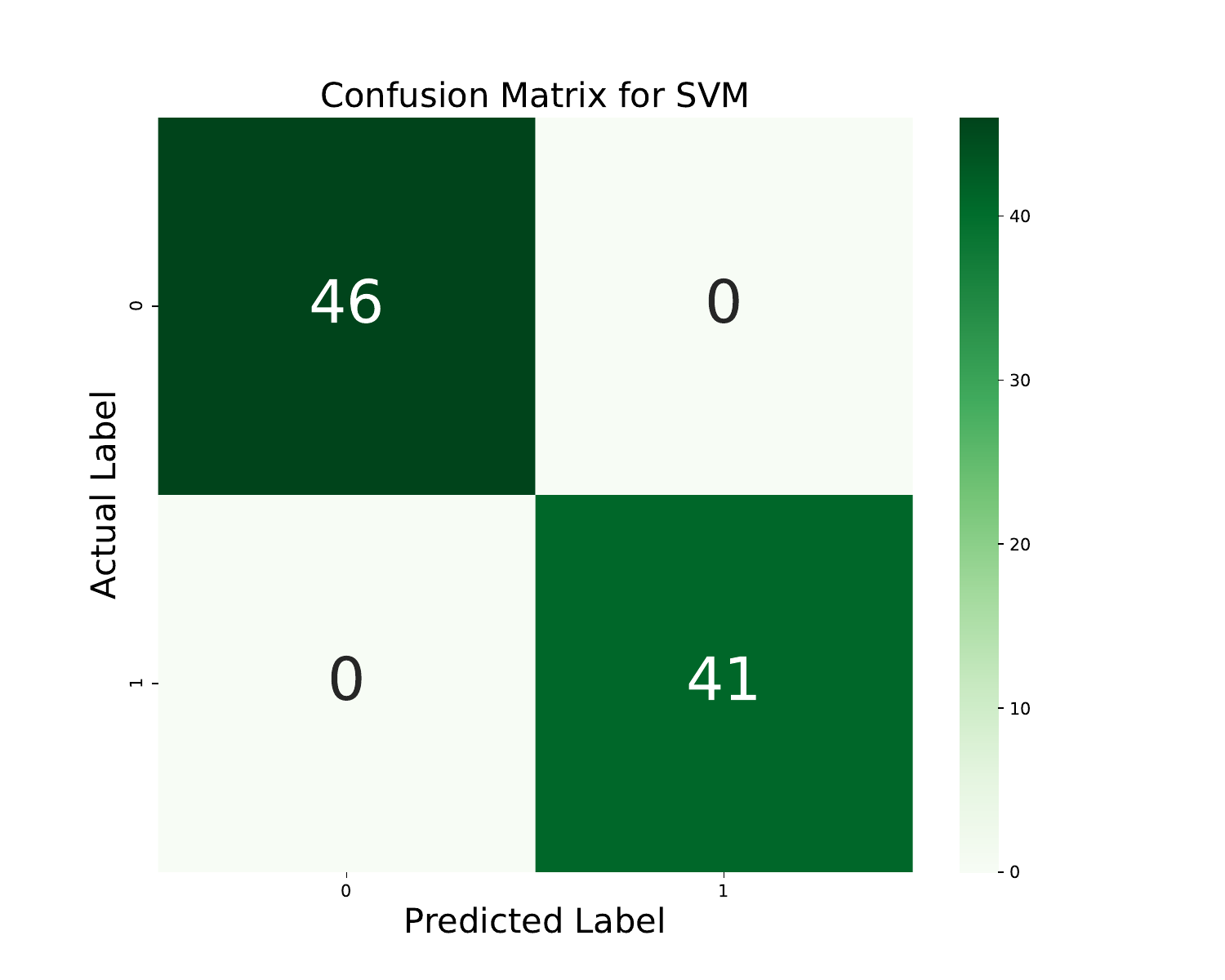}
    \includegraphics[width=.24\textwidth]{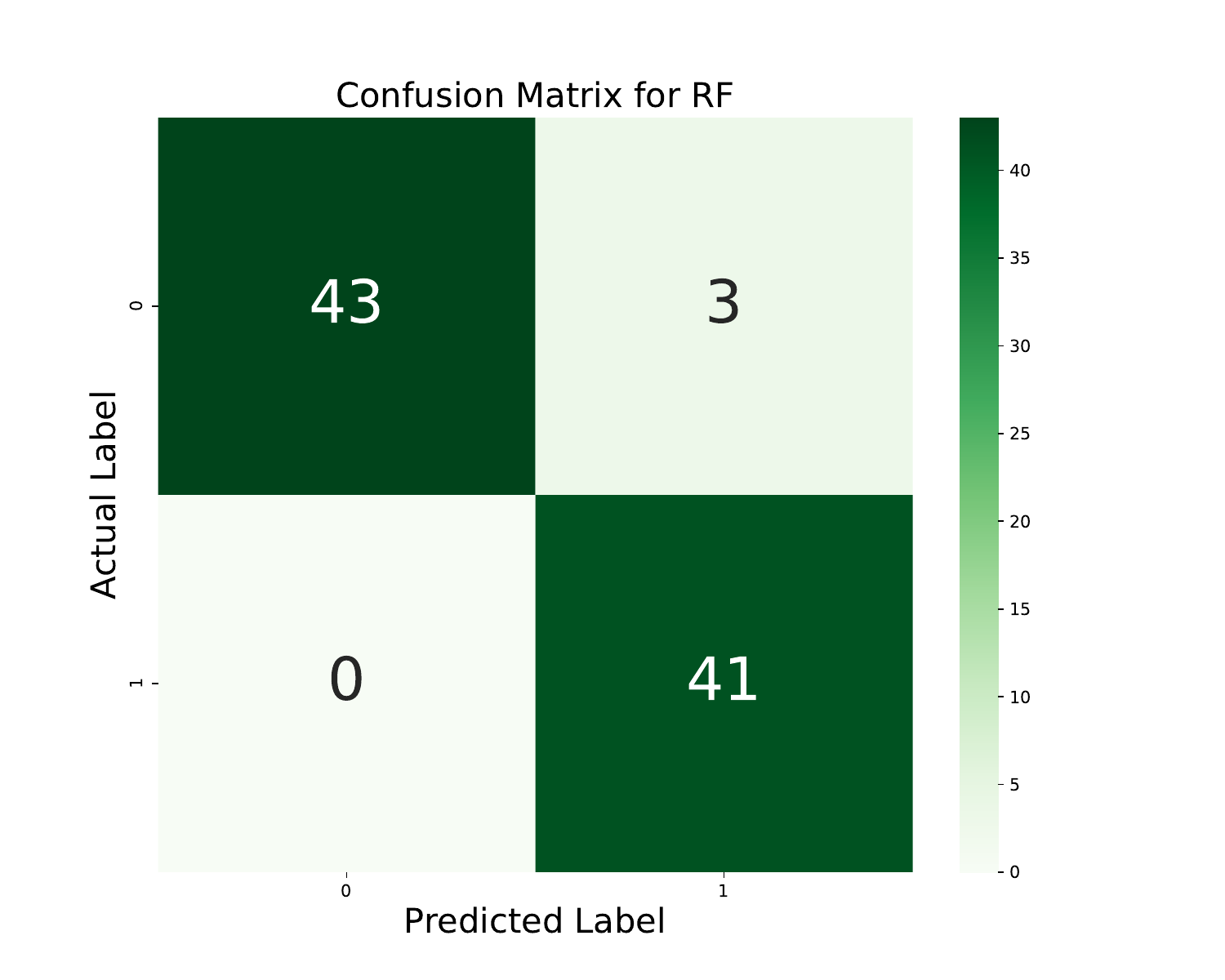}
\hfill

    \includegraphics[width=.24\textwidth]{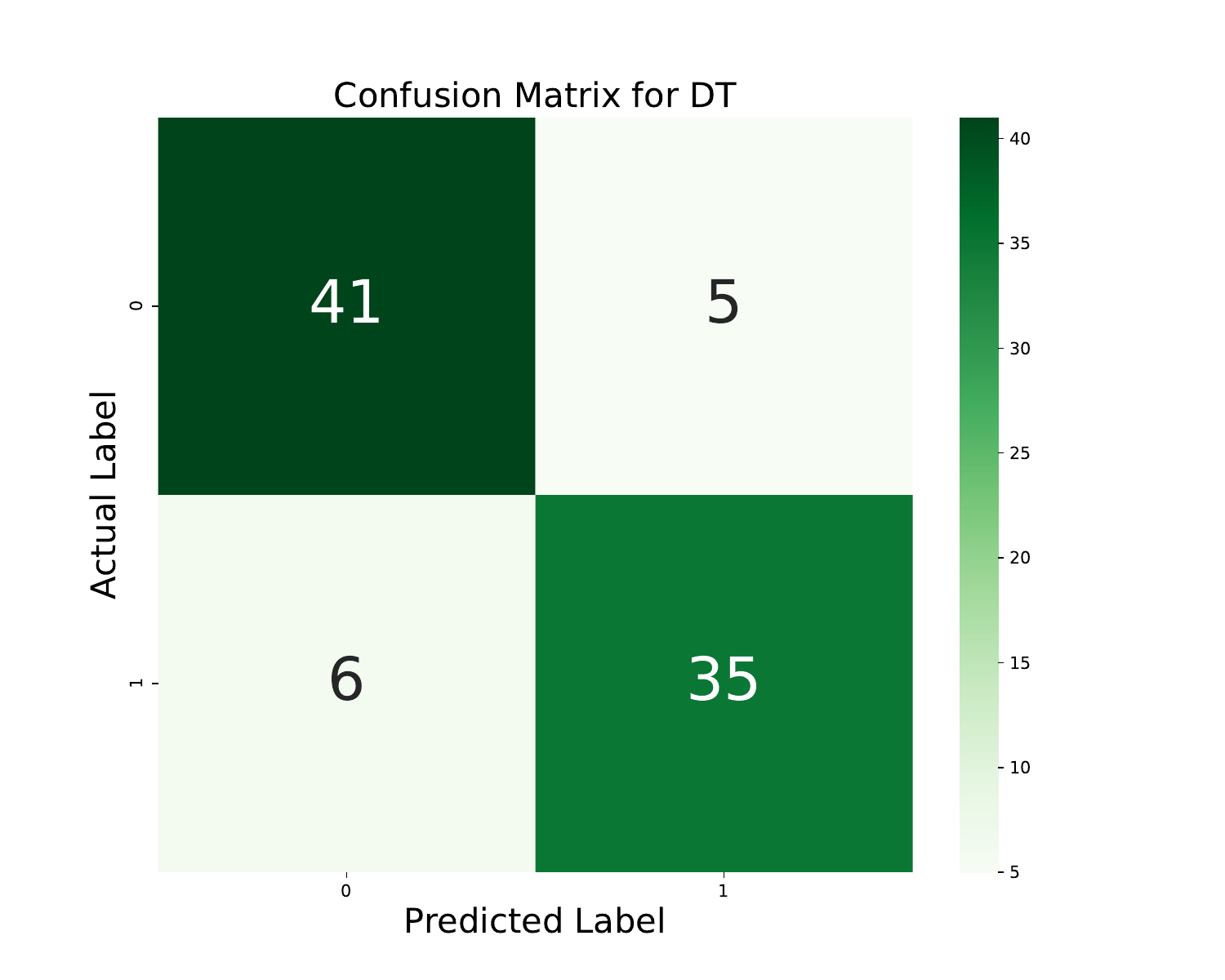}
    \includegraphics[width=.24\textwidth]{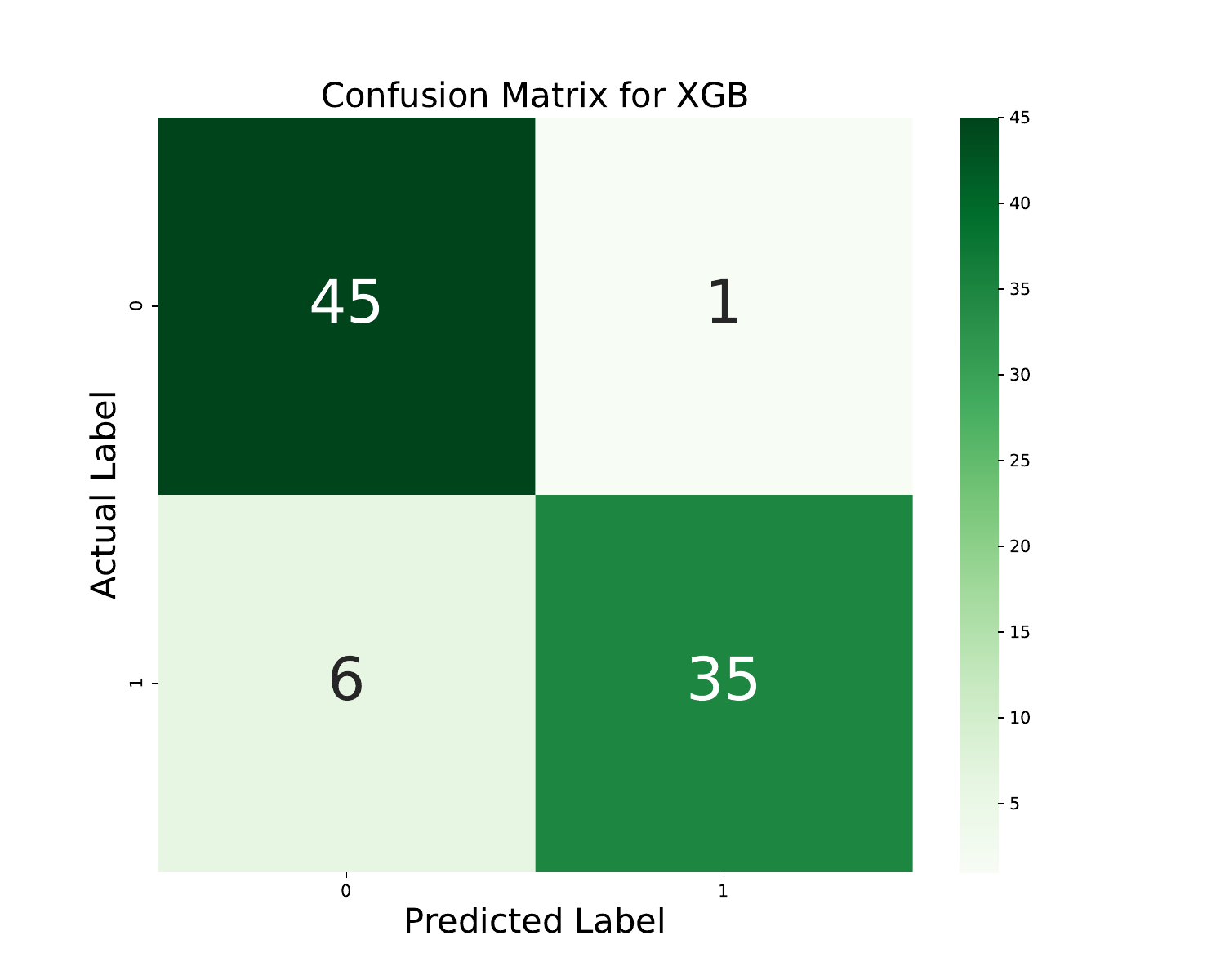}
    \includegraphics[width=.24\textwidth]{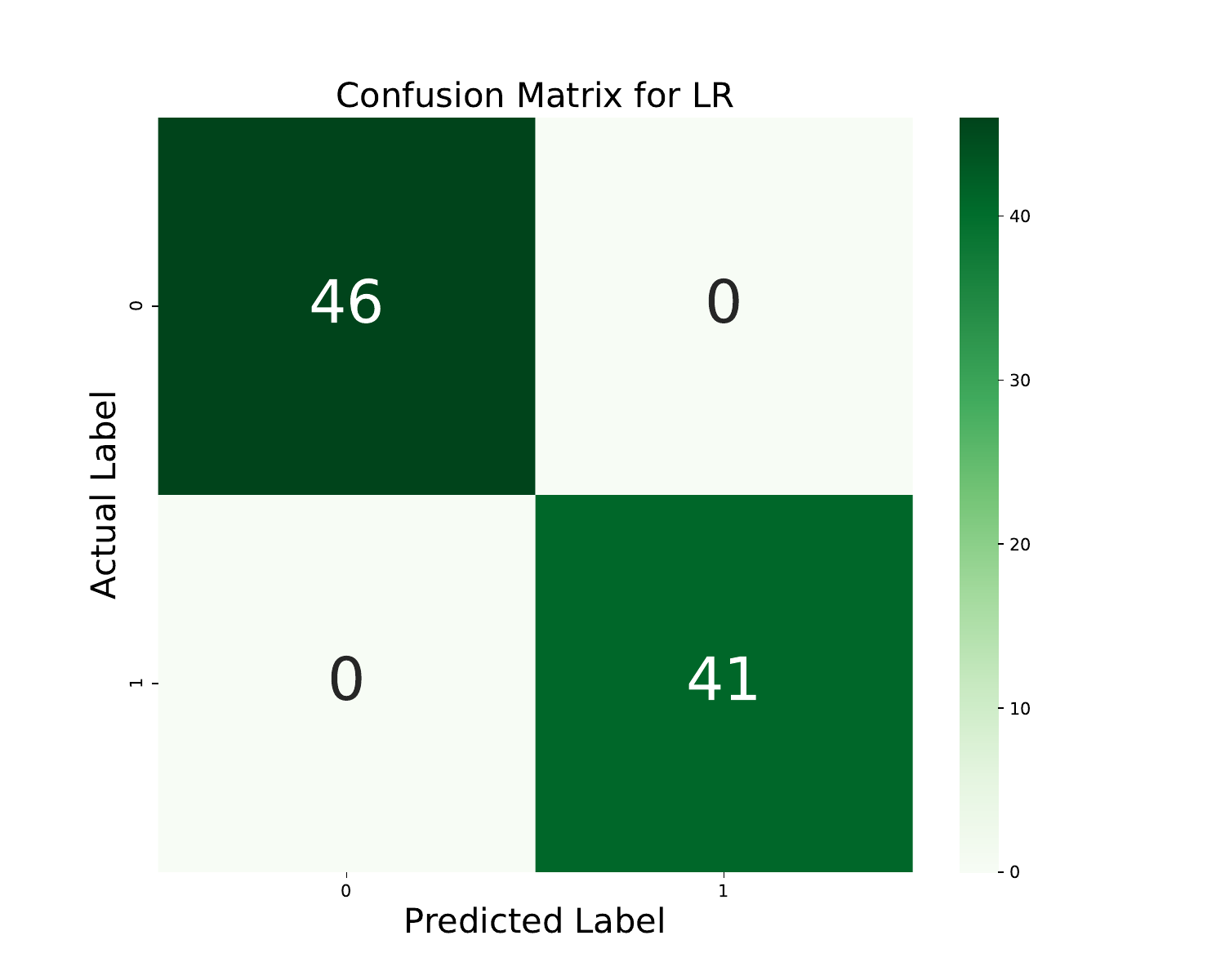}
    \includegraphics[width=.24\textwidth]{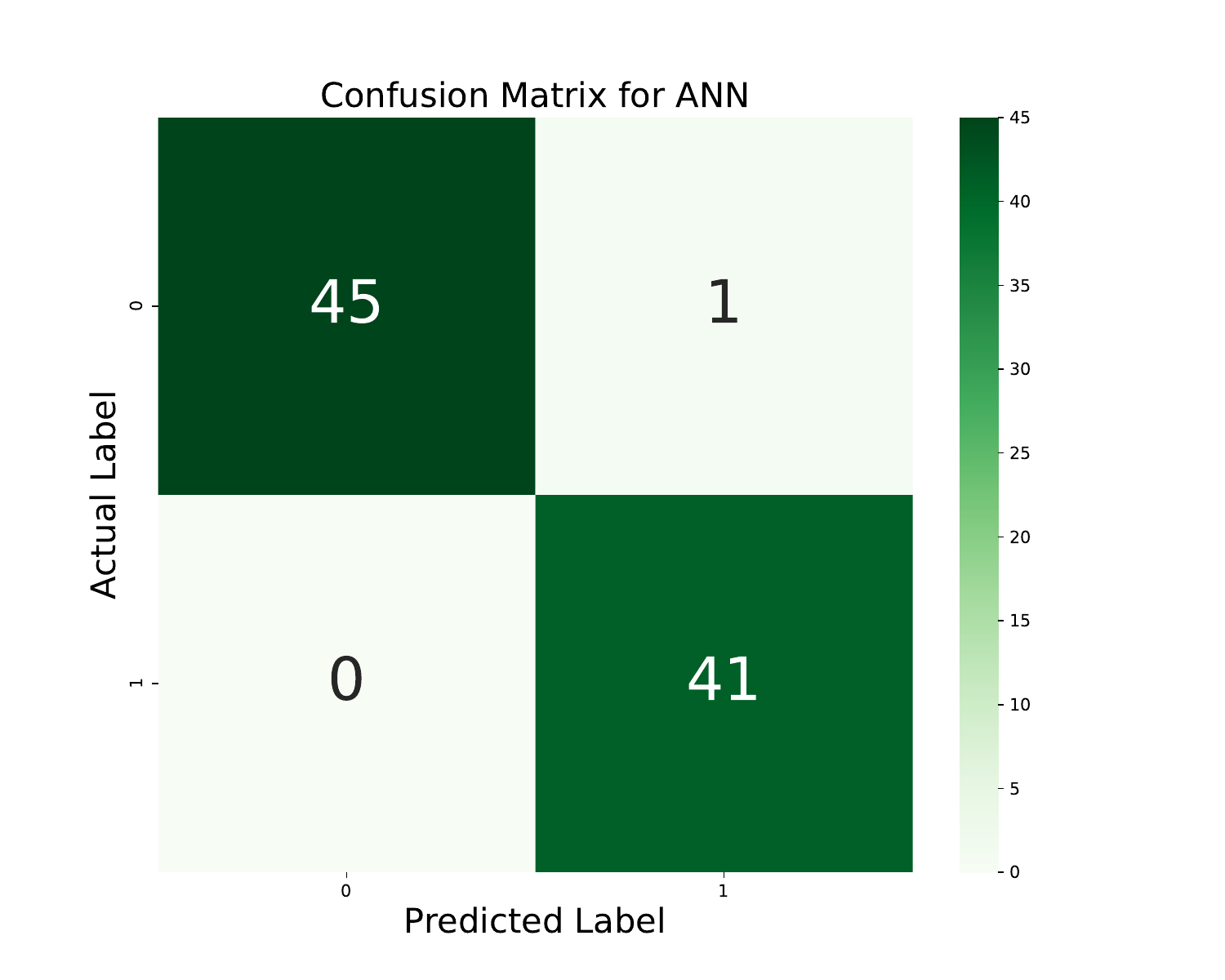}
\hfill

\caption{All Confusion matrix for Child dataset}
\label{fig:child_conf}
\end{figure*}

\begin{figure*}
\centering
    \includegraphics[width=.24\textwidth]{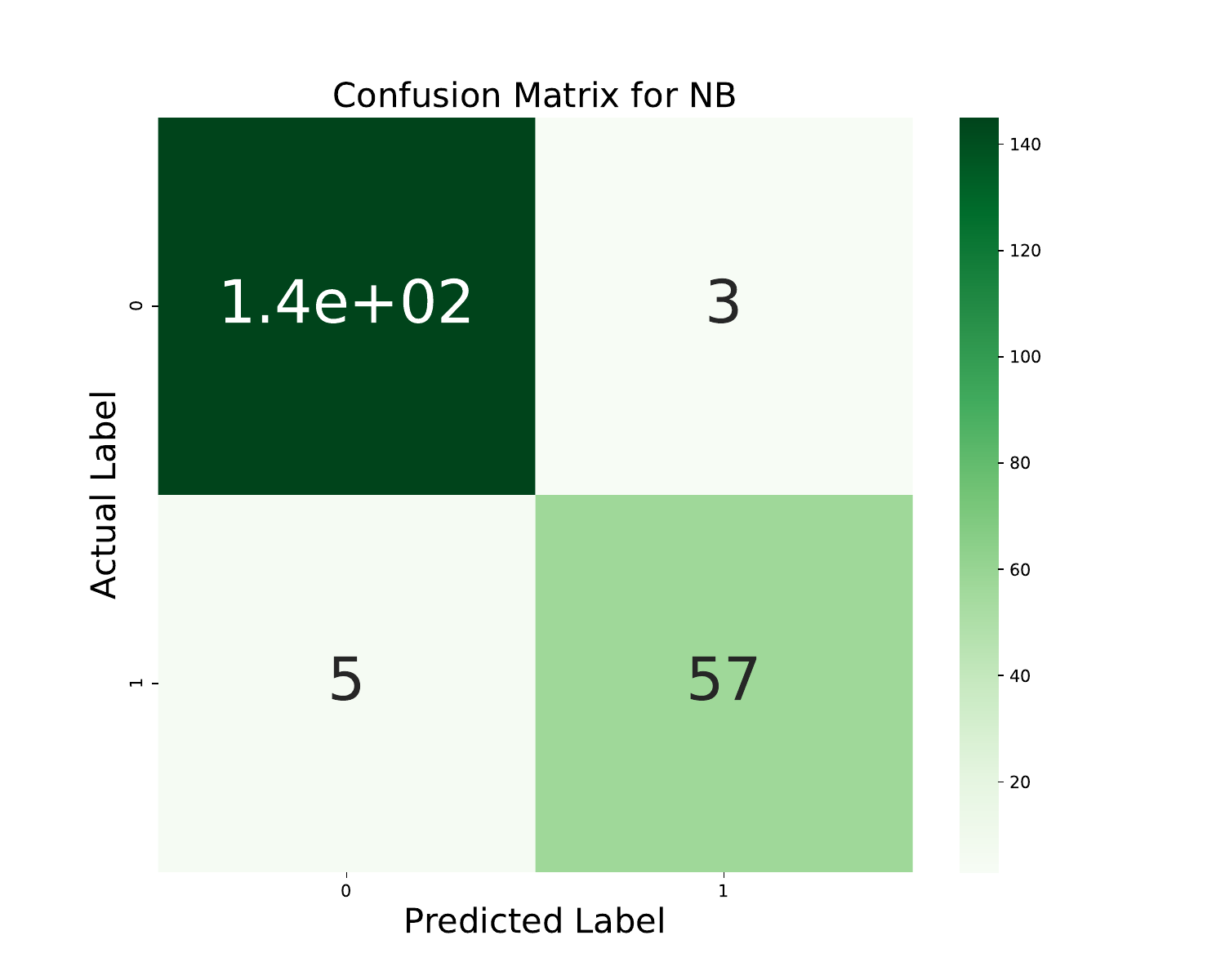}
    \includegraphics[width=.24\textwidth]{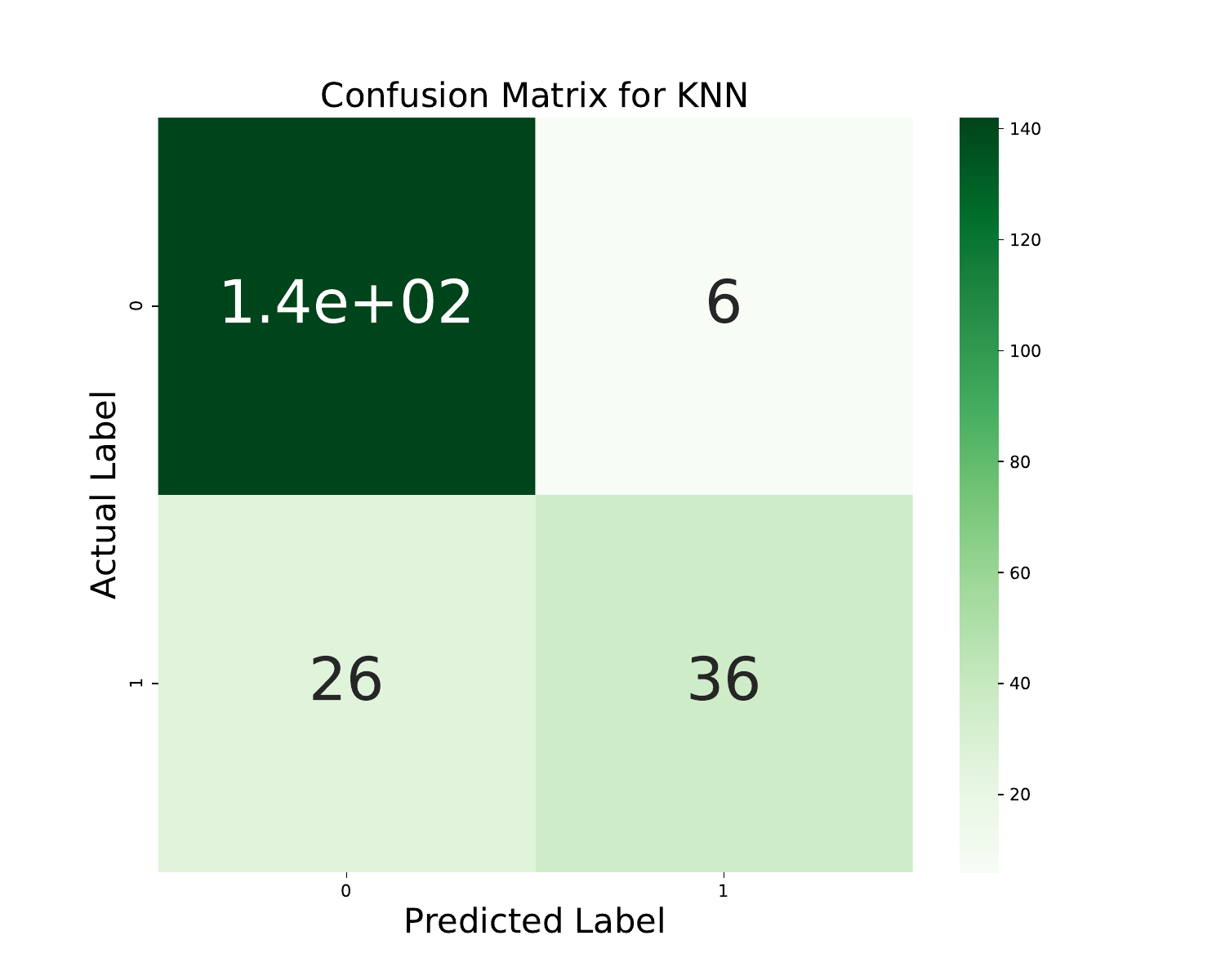}
    \includegraphics[width=.24\textwidth]{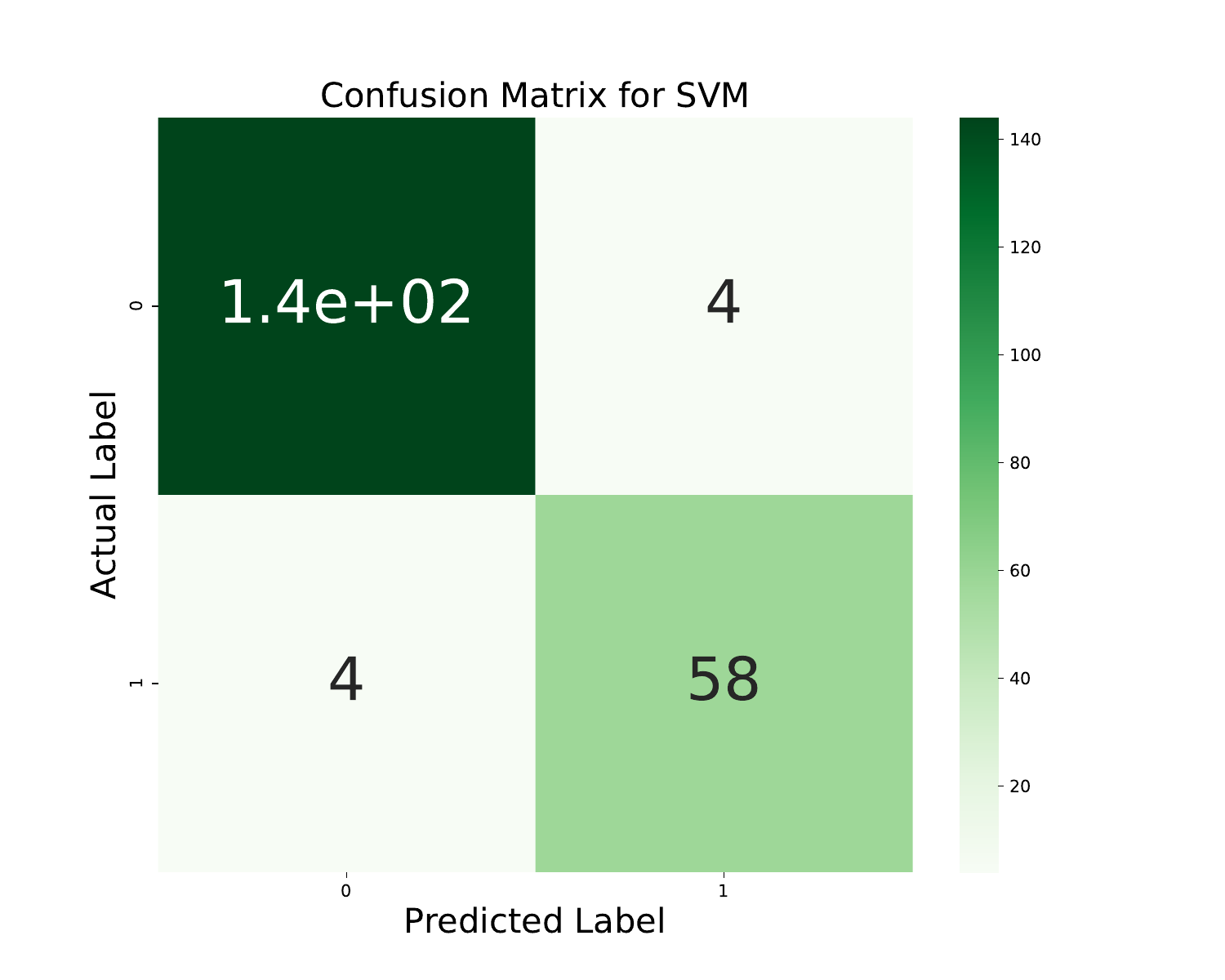}
    \includegraphics[width=.24\textwidth]{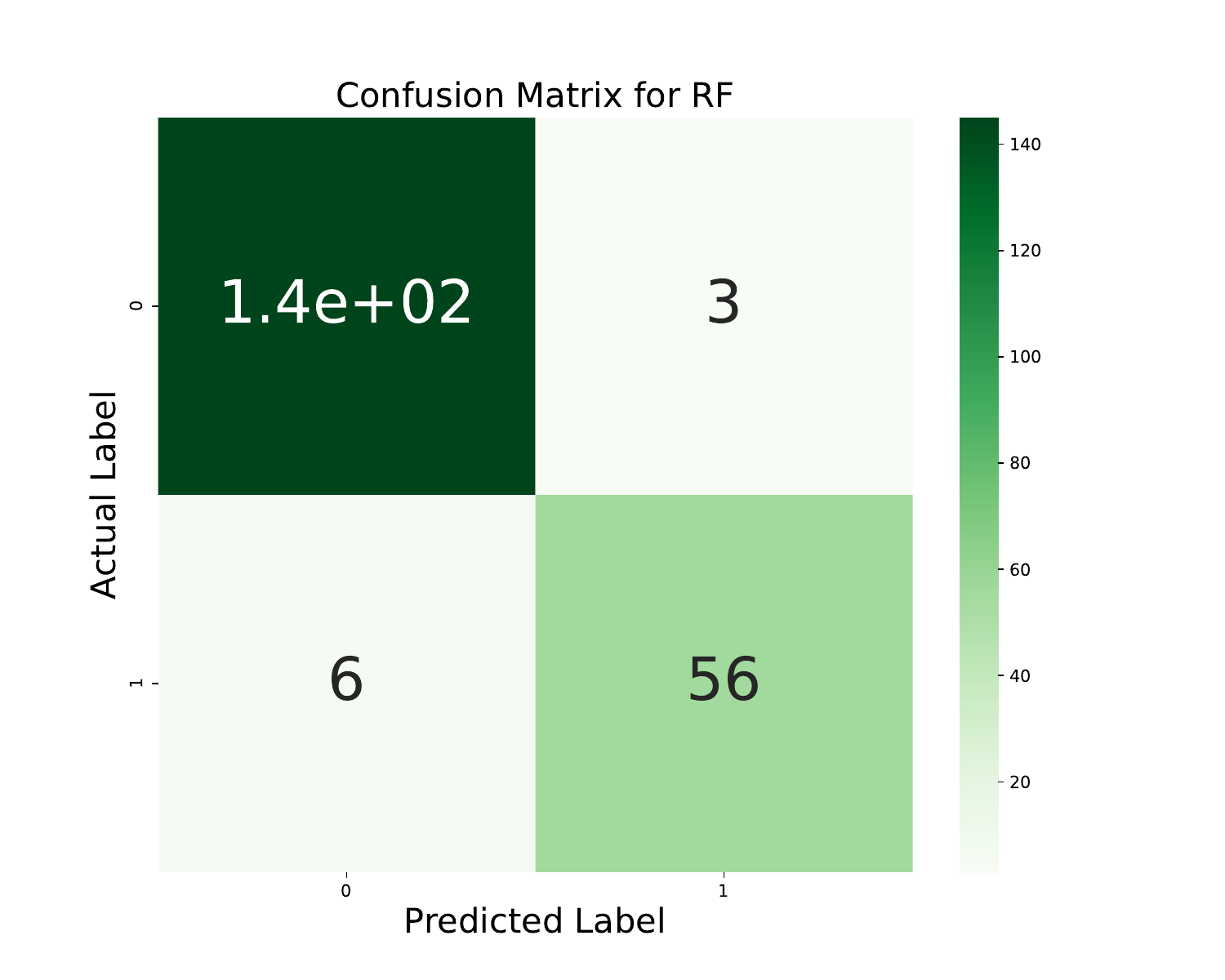}
\hfill

    \includegraphics[width=.24\textwidth]{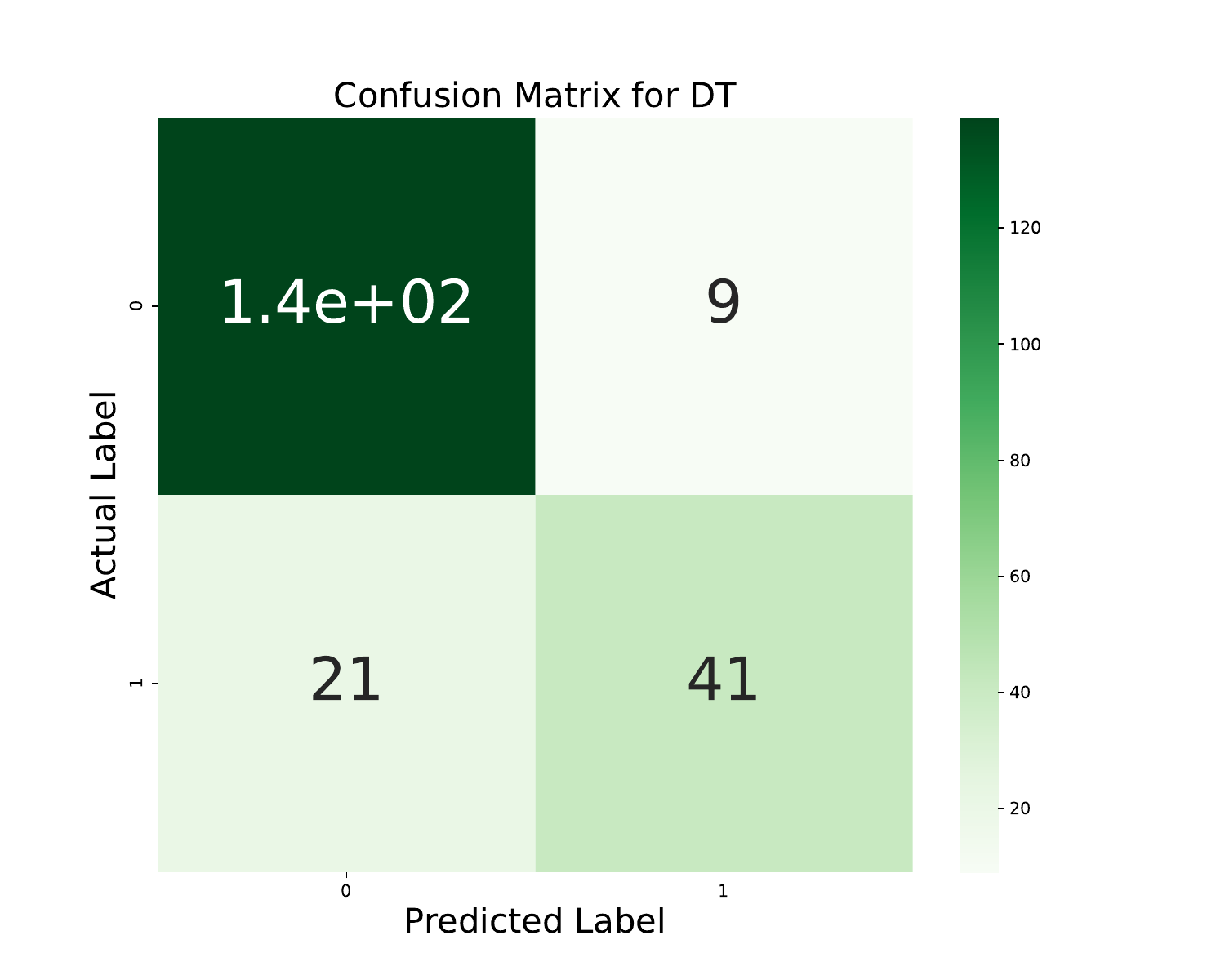}
    \includegraphics[width=.24\textwidth]{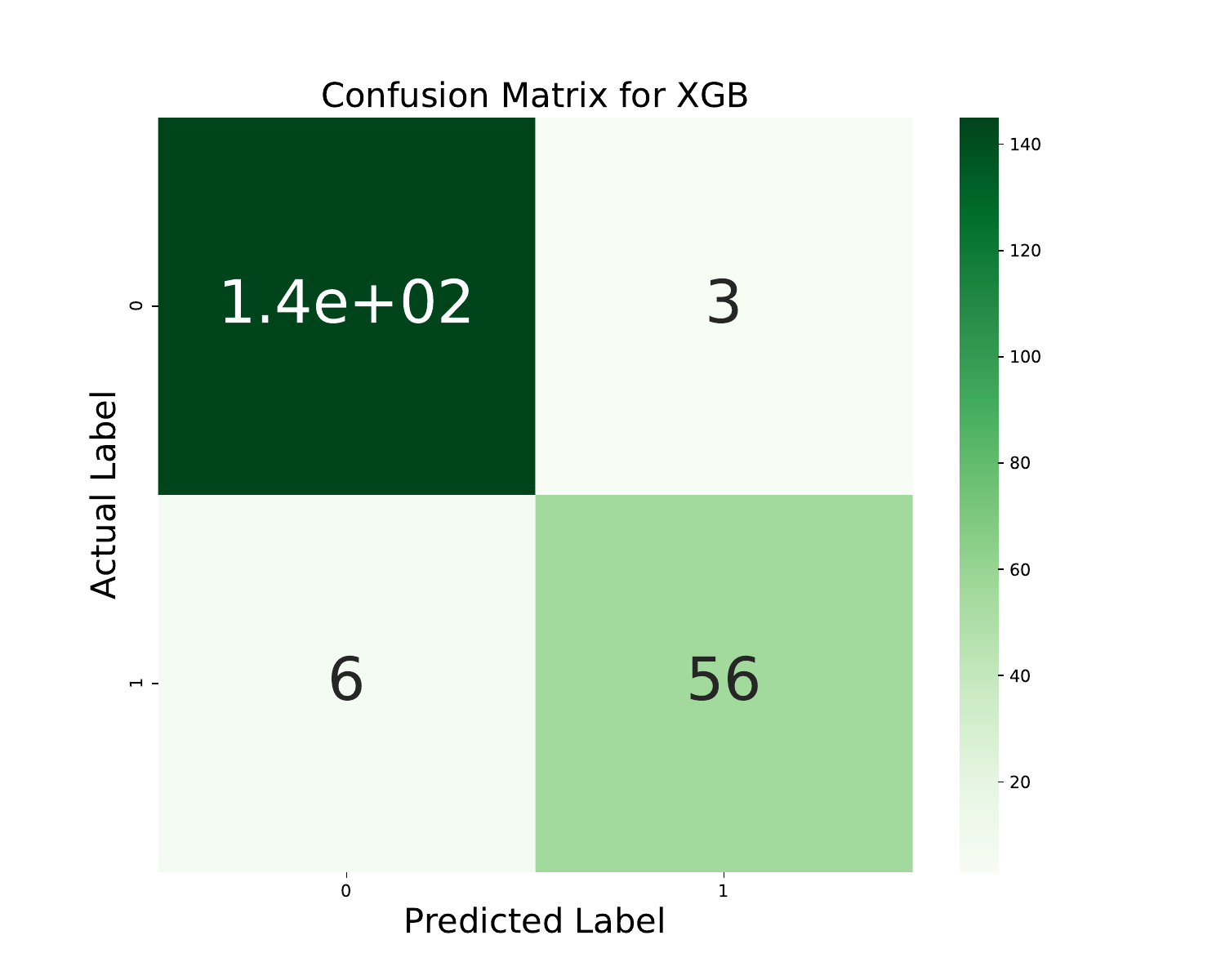}
    \includegraphics[width=.24\textwidth]{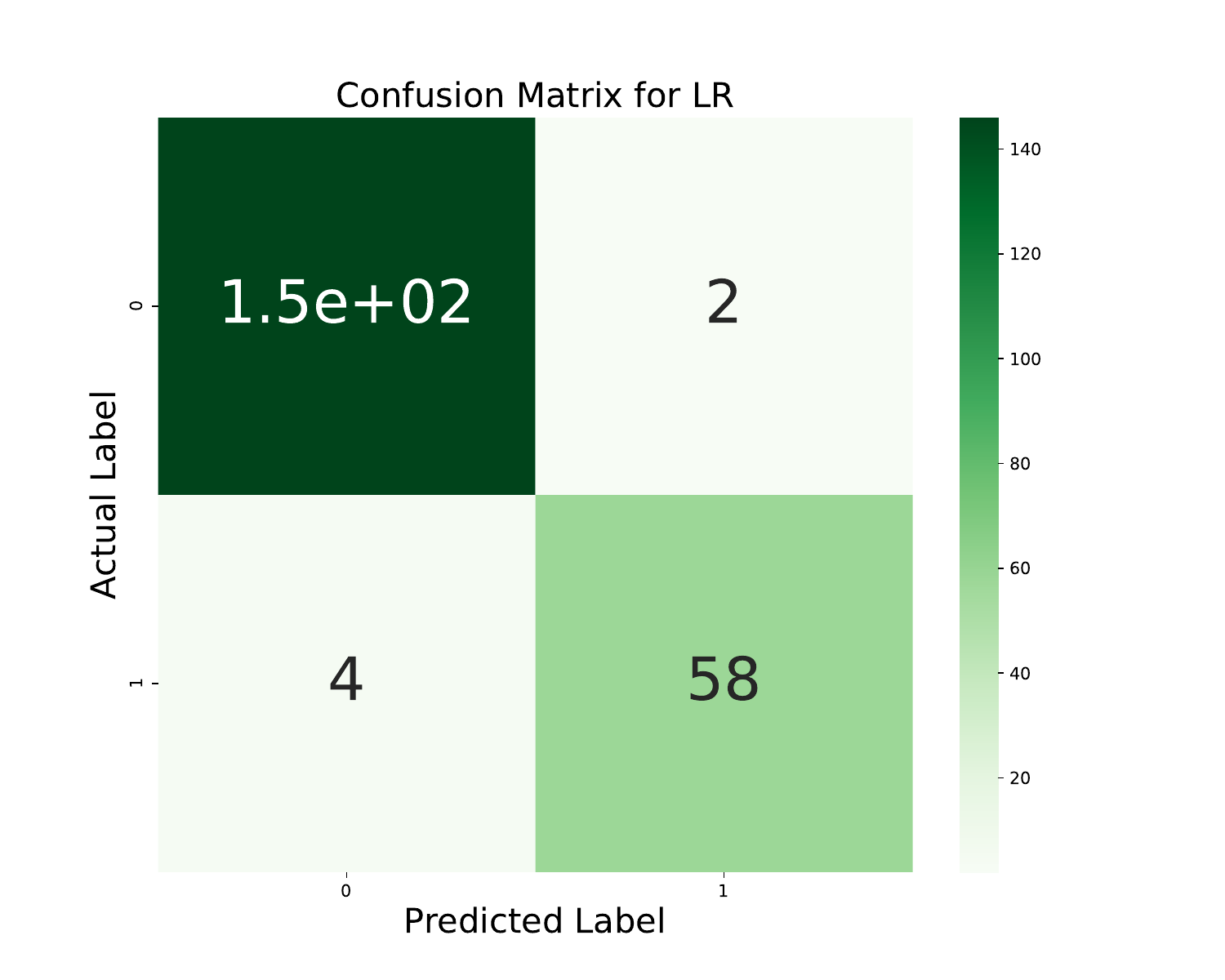}
    \includegraphics[width=.24\textwidth]{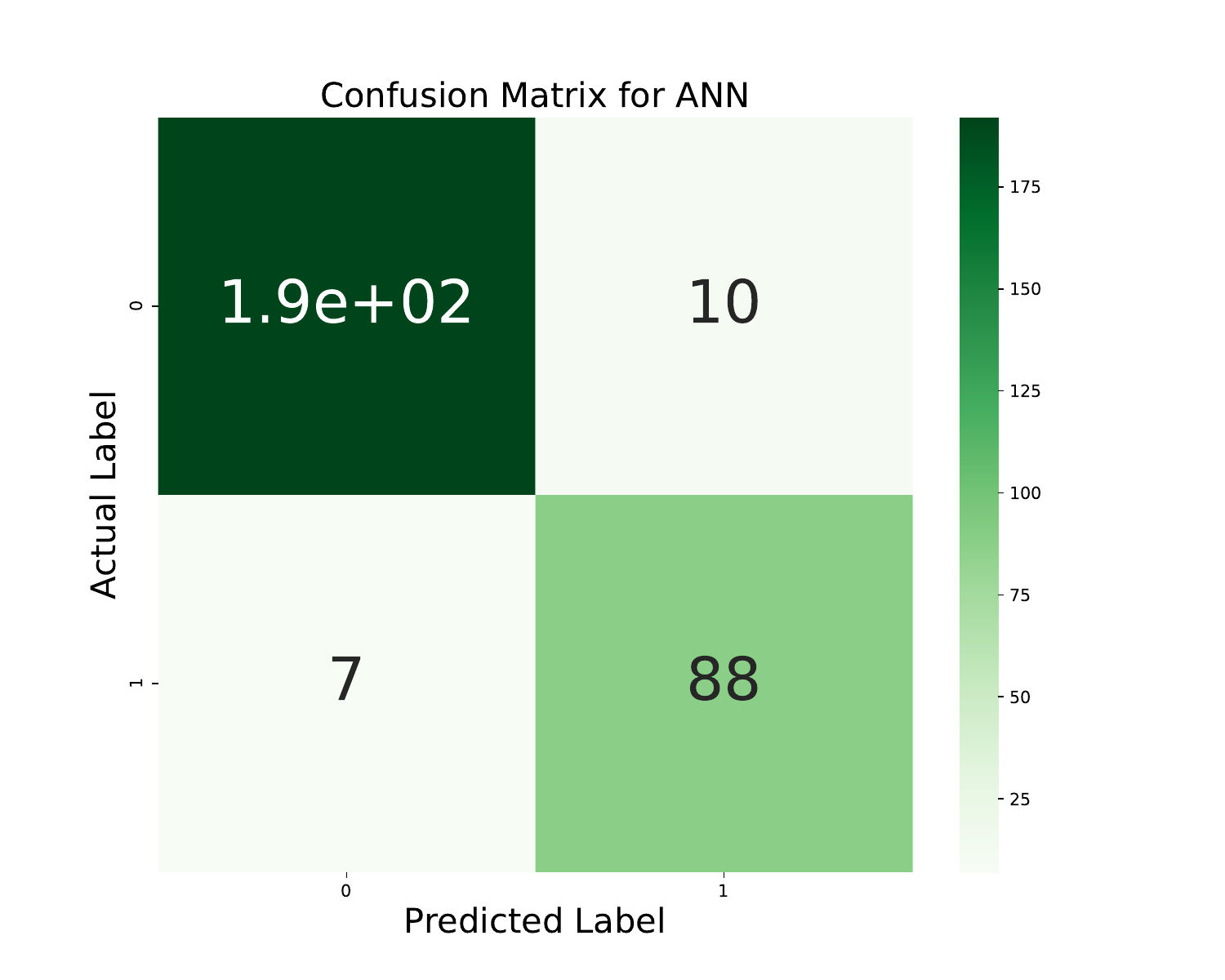}
\hfill

\caption{All Confusion matrix for Adult dataset}
\label{fig:adult_conf}
\end{figure*}

\begin{figure*}
\centering
    \includegraphics[width=.24\textwidth]{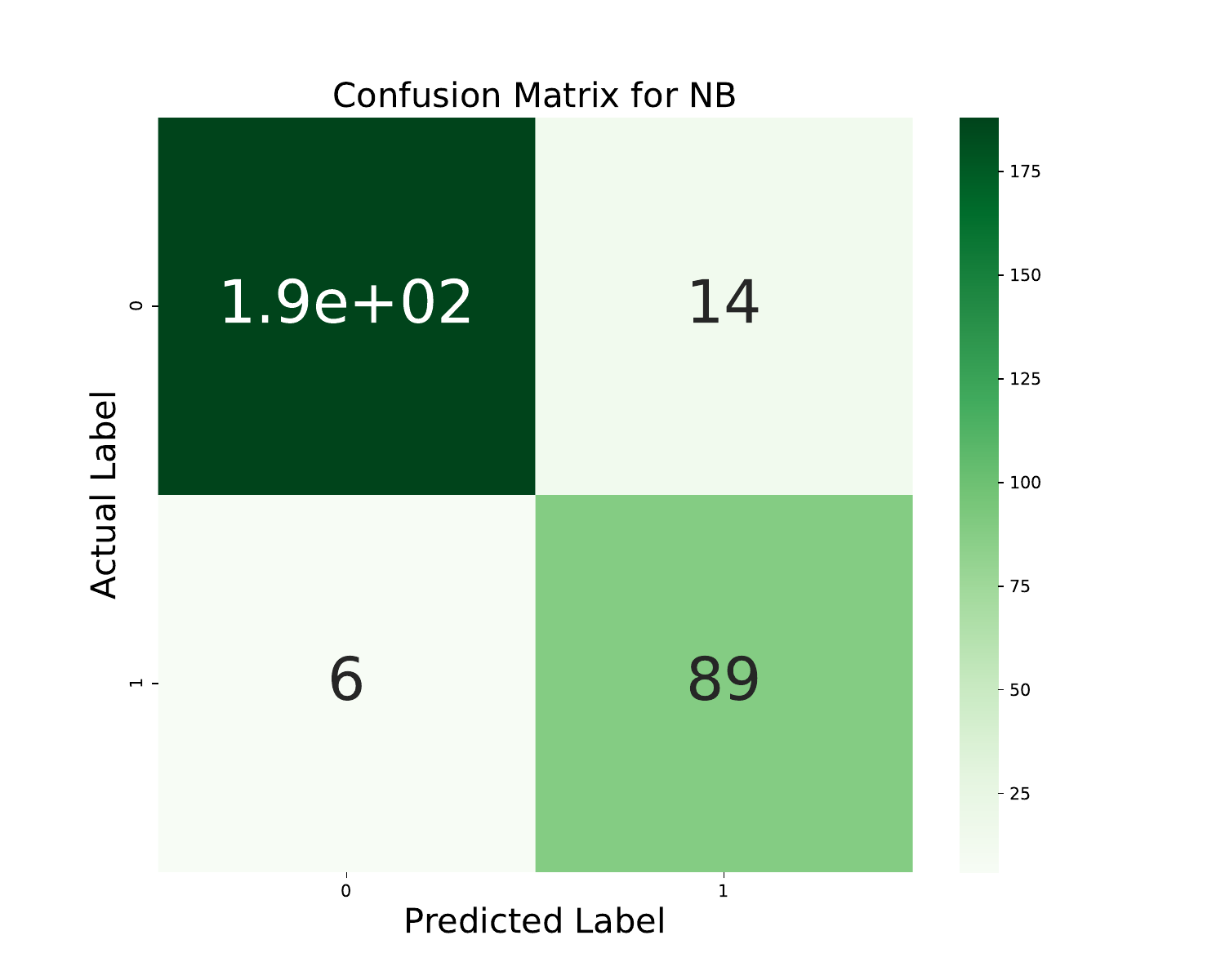}
    \includegraphics[width=.24\textwidth]{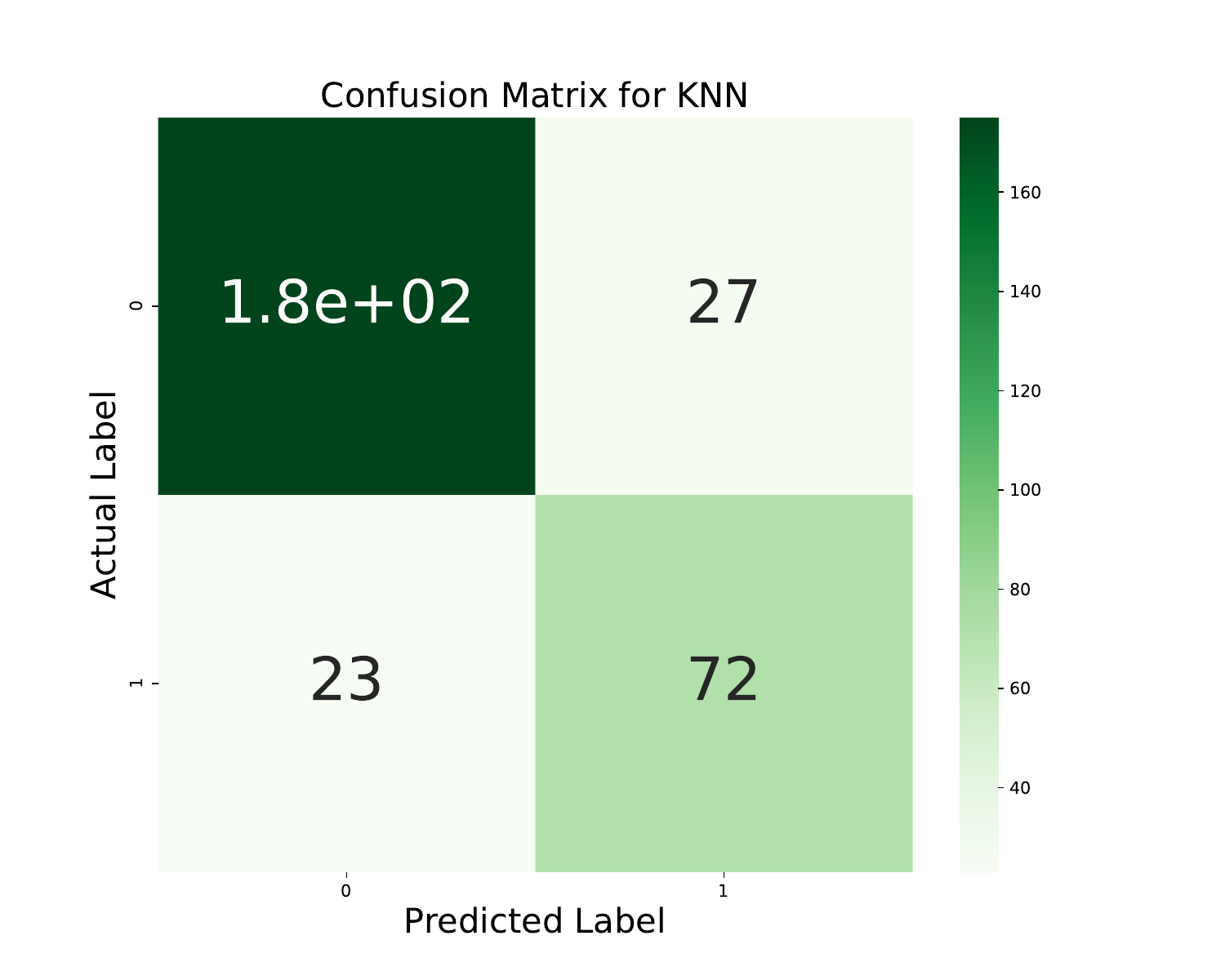}
    \includegraphics[width=.24\textwidth]{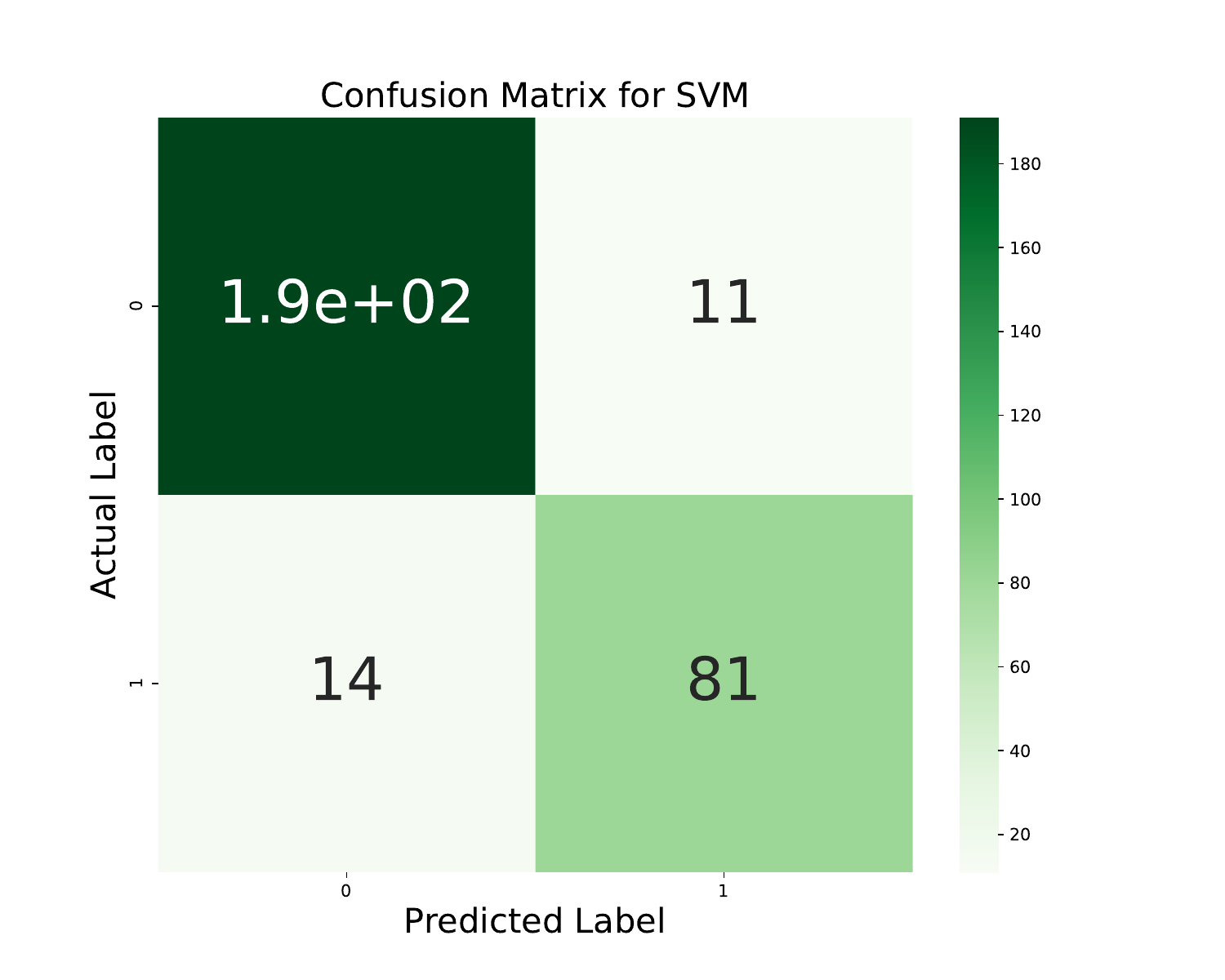}
    \includegraphics[width=.24\textwidth]{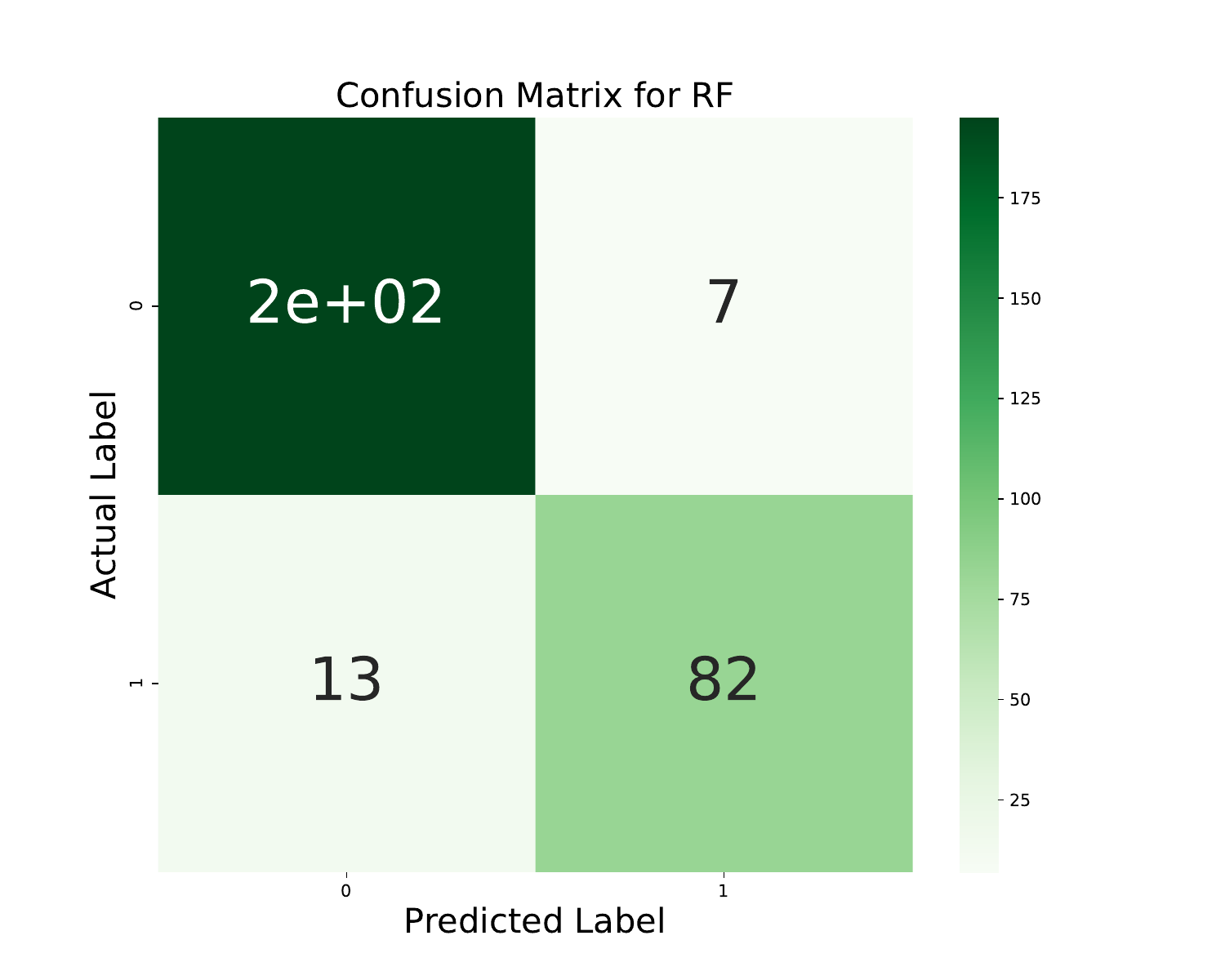}
\hfill

    \includegraphics[width=.24\textwidth]{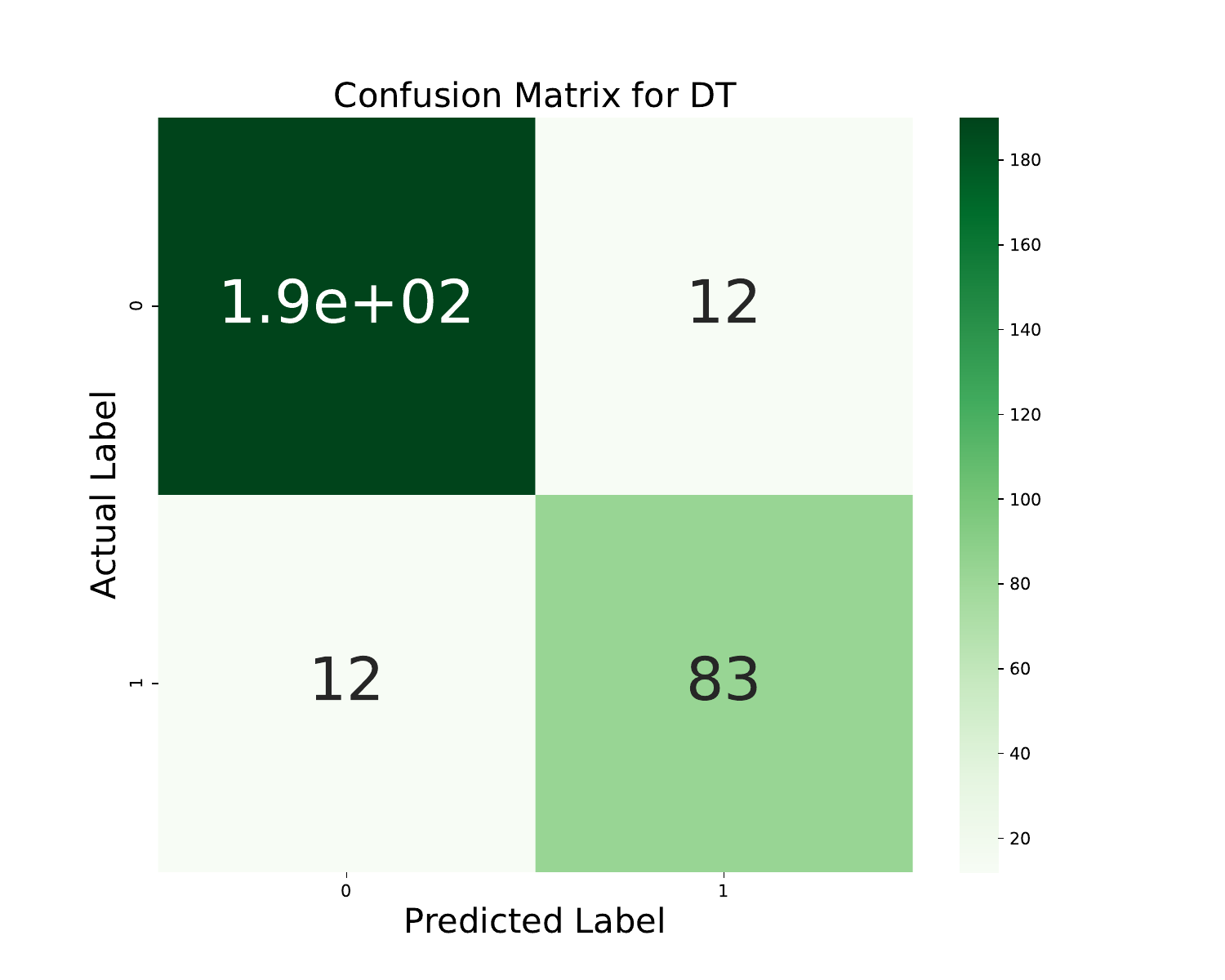}
    \includegraphics[width=.24\textwidth]{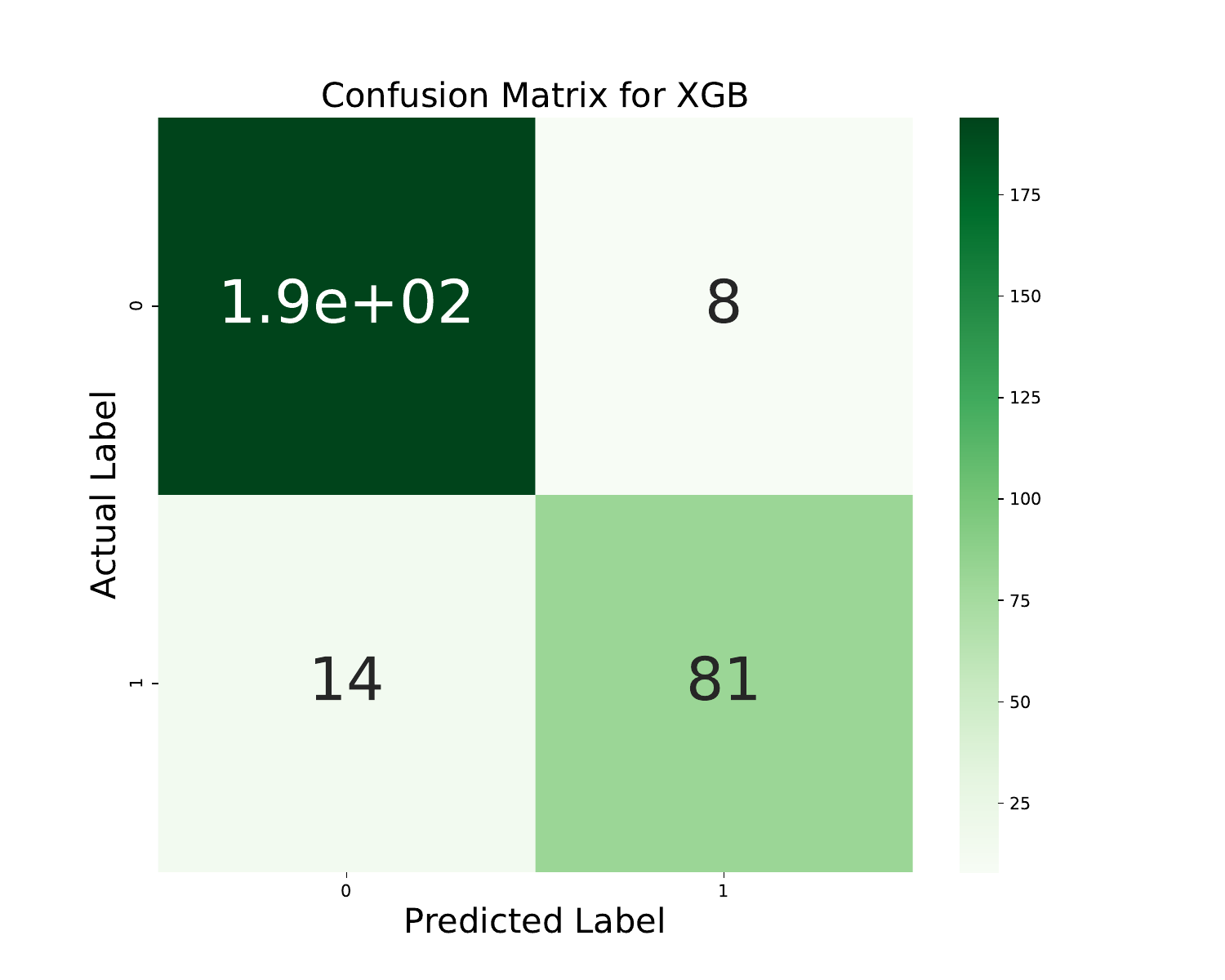}
    \includegraphics[width=.24\textwidth]{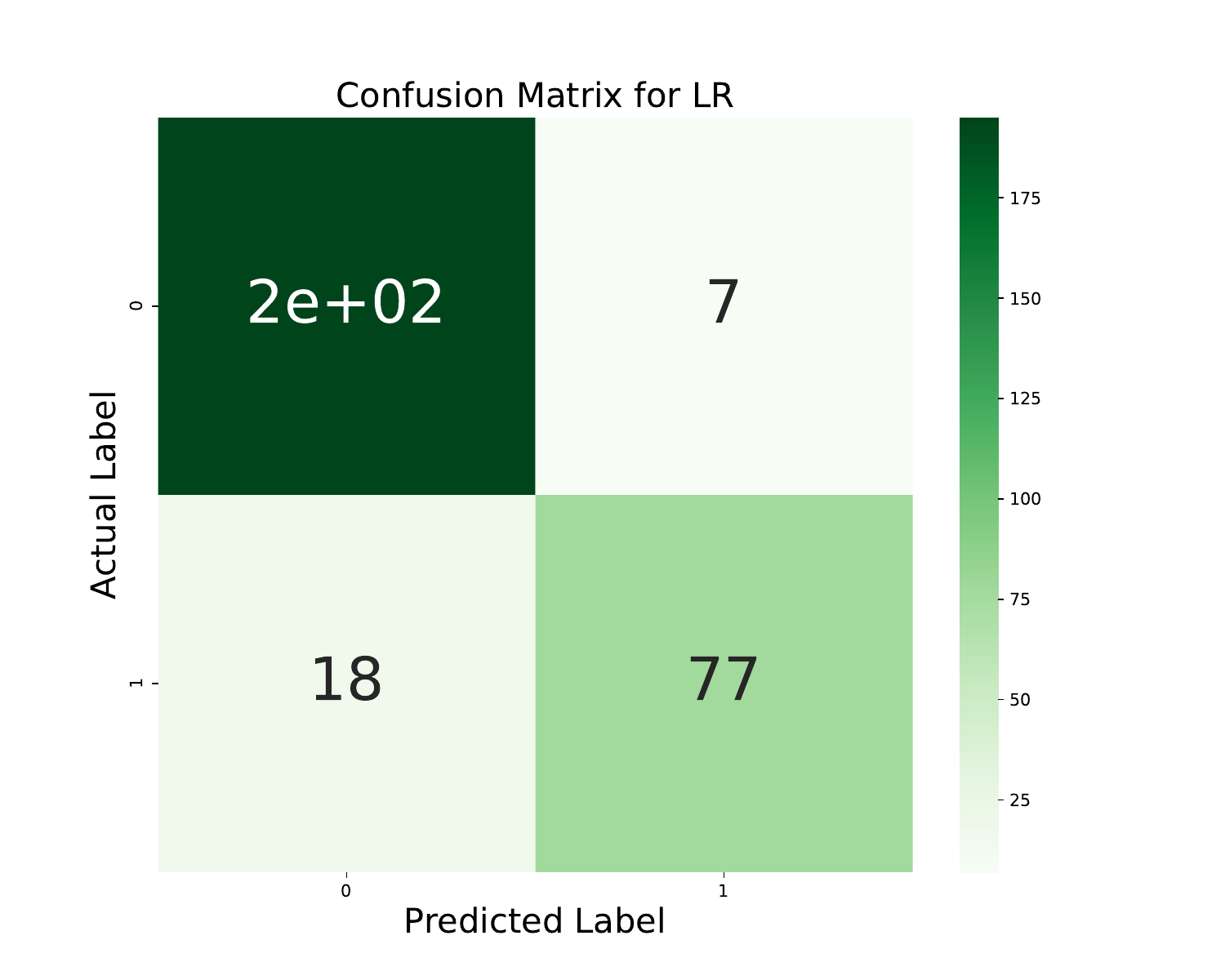}
    \includegraphics[width=.24\textwidth]{MLP_confusion_cb.pdf}
\hfill

\caption{All Confusion matrix for Combined dataset}
\label{fig:comb_conf}
\end{figure*}

From the Table \ref{tab:child}, we see that the Support Vector Machine and Logistic Regression models are provided the 100\% meteric values in terms of all evaluation metrics and 0.00 in log loss value. The Random Forests and Artificial Neural Network models are also shown 100\% in Recall metrics. If we look at the Table \ref{tab:distibution} and Figure \ref{fig:distribution} then we can easily understand that the number of instances of two classes is not highly imbalanced. That's why machine learning models can easily produce higher accuracy by utilizing their best hyperparameters. On the other hand the other two datasets are highly imbalanced. 

Table \ref{tab:adult} shows that the Logistic Regression model provides the highest metrics values of 97.14\%, 96.67\%, 98.65\%, 95.08\%, and 93.07\% in accuracy, precision, specificity, f1-score, and kappa-score, respectively, and the lowest log loss value of 1.029. The ANN model gives the highest values of 98.39\% and 97.17\% in recall and AUC, respectively. In the adult dataset, we found that the dataset is highly imbalanced. In machine learning studies, the class imbalance problem affects the models, and due to this, the model cannot be classified efficiently.

In this study, we have generated a new combined dataset by using the children and adult datasets to investigate how our utilized ML models are performed. In addition, we take advantage of the benefits of a new dataset that includes social screening data from both children and adults. This way, we will not need different datasets. The ML models can easily identify the ASD from one dataset for both children and adults. The ML model performance for the new combined dataset is tabulated in Table \ref{tab:combined}. From Table \ref{tab:combined}, it is seen that the ANN model provides the best accuracy, f1-score, AUC, kappa, and loss; RF is given the precision and specificity; and NB provides the best recall value. 

\begin{figure*}[]
\centering

    \includegraphics[width=.49\textwidth]{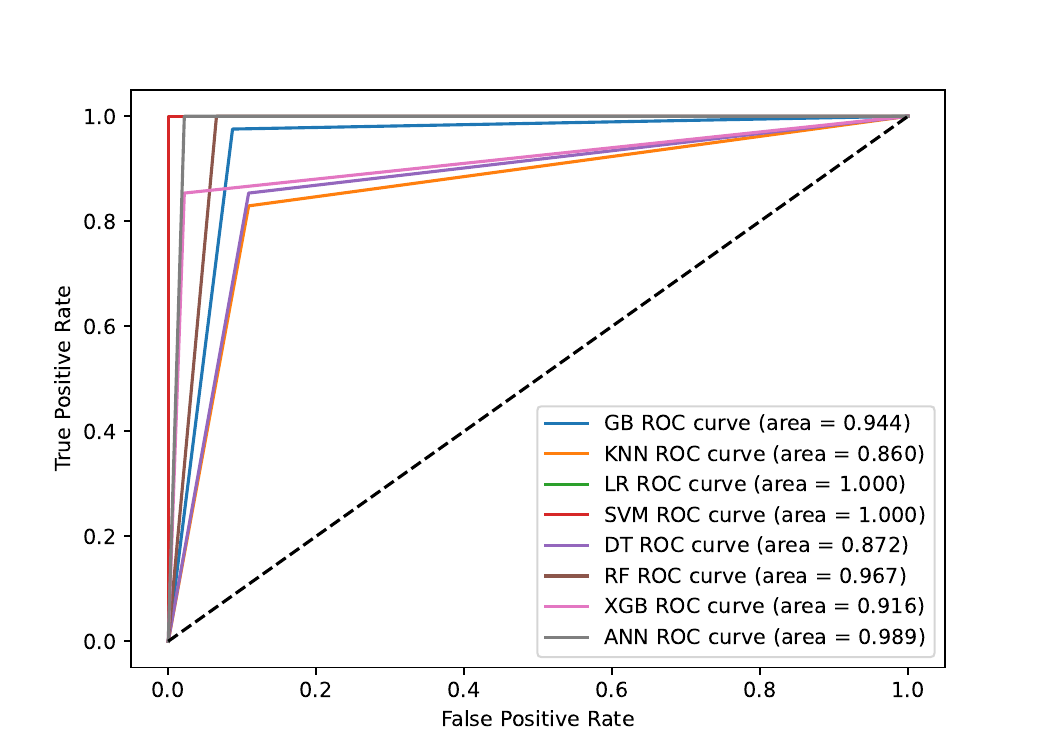}
    \includegraphics[width=.49\textwidth]{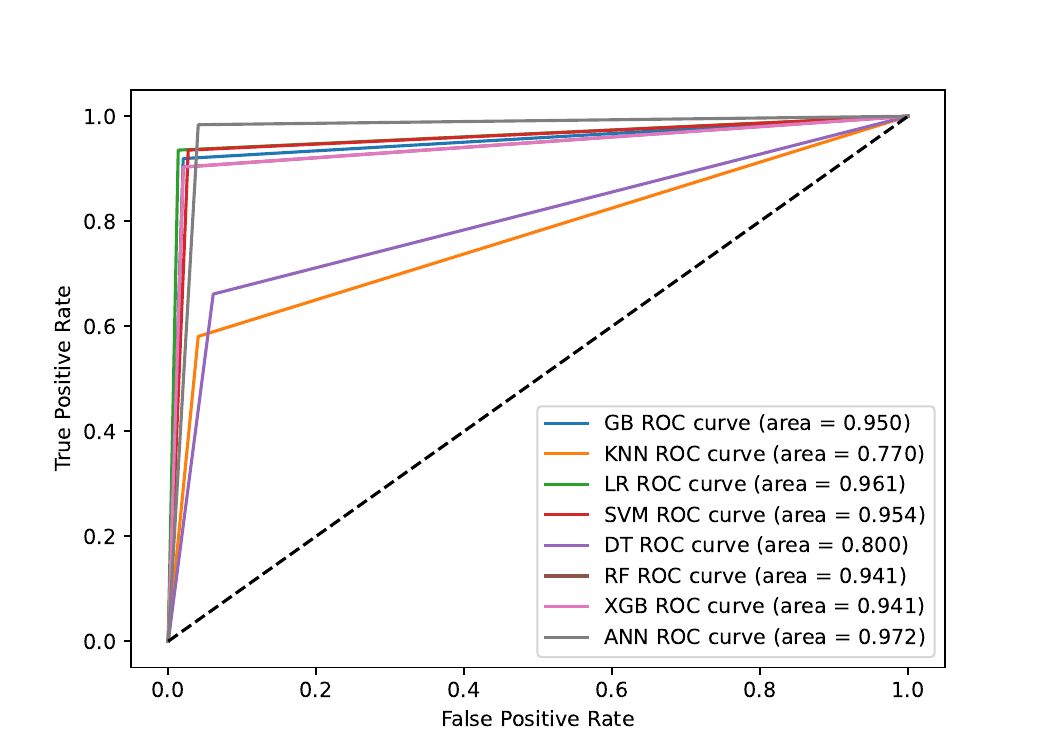}
    \includegraphics[width=.49\textwidth]{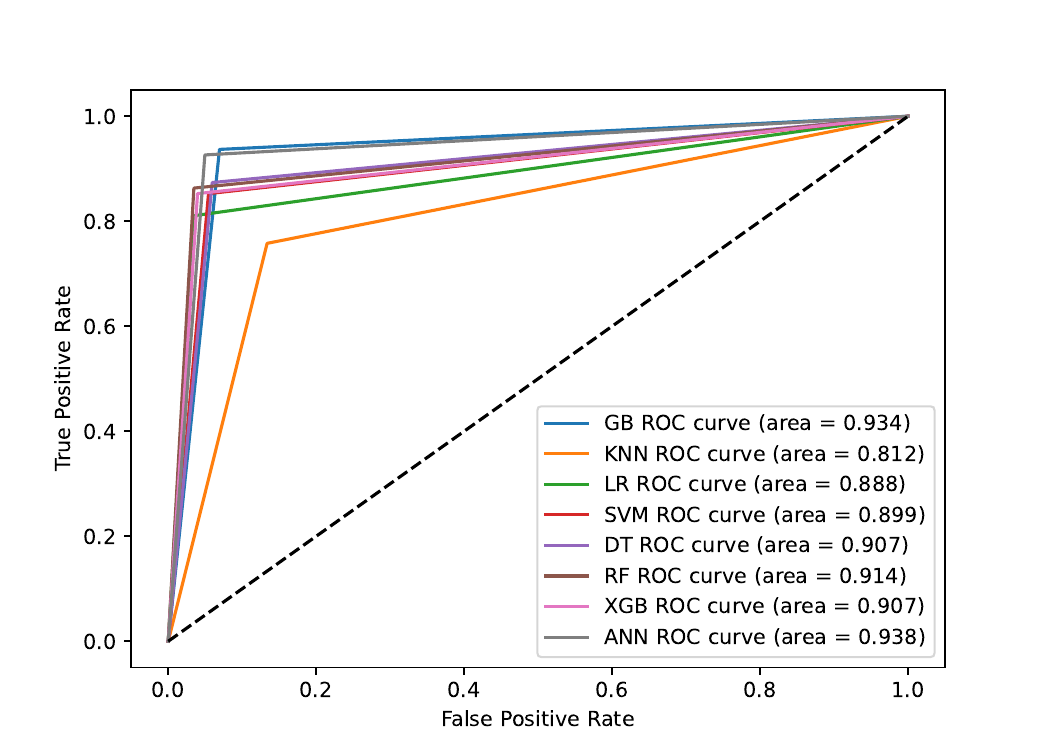}

\caption{AUC-ROC curve comparison for Child, Adult and Combined dataset}
\label{fig:roc}
\end{figure*}

We also added the bar charts based on the Table \ref{tab:child}, Table \ref{tab:adult}, and Table \ref{tab:combined} values for the different datasets because one can understand the comparison the metrics result from the graphical representations. The confusion matrices are also added in Figures \ref{fig:child_conf}, \ref{fig:adult_conf}, and \ref{fig:comb_conf} for the different datasets to understand a clear image in all the evaluation metrics. The ROC-AUC curve comparison figure is given in Figure \ref{fig:roc}. The model's efficiency is shown by the ROC AUC score.

\subsection{Clustering results}
\begin{figure*}[]
\centering

    \includegraphics[width=.49\textwidth]{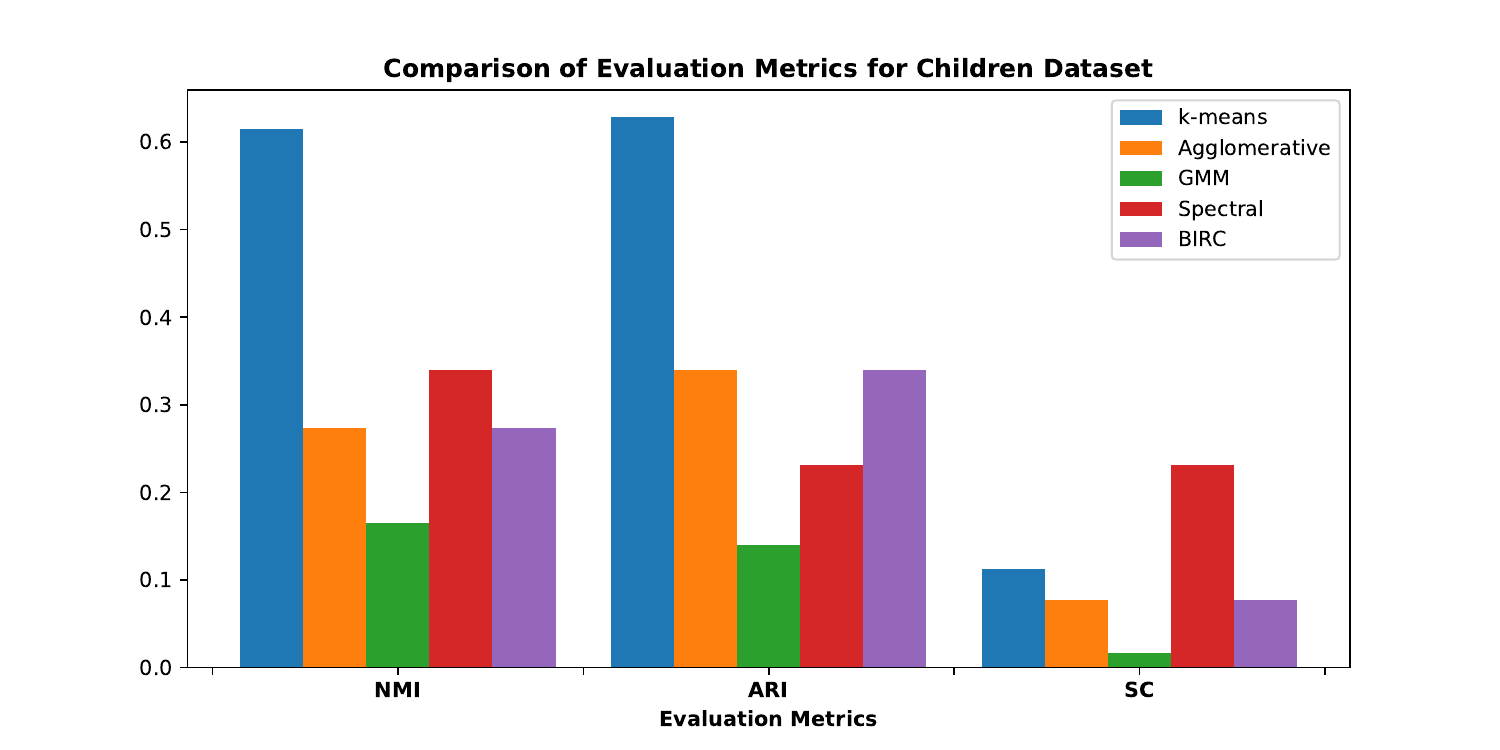}
    \includegraphics[width=.49\textwidth]{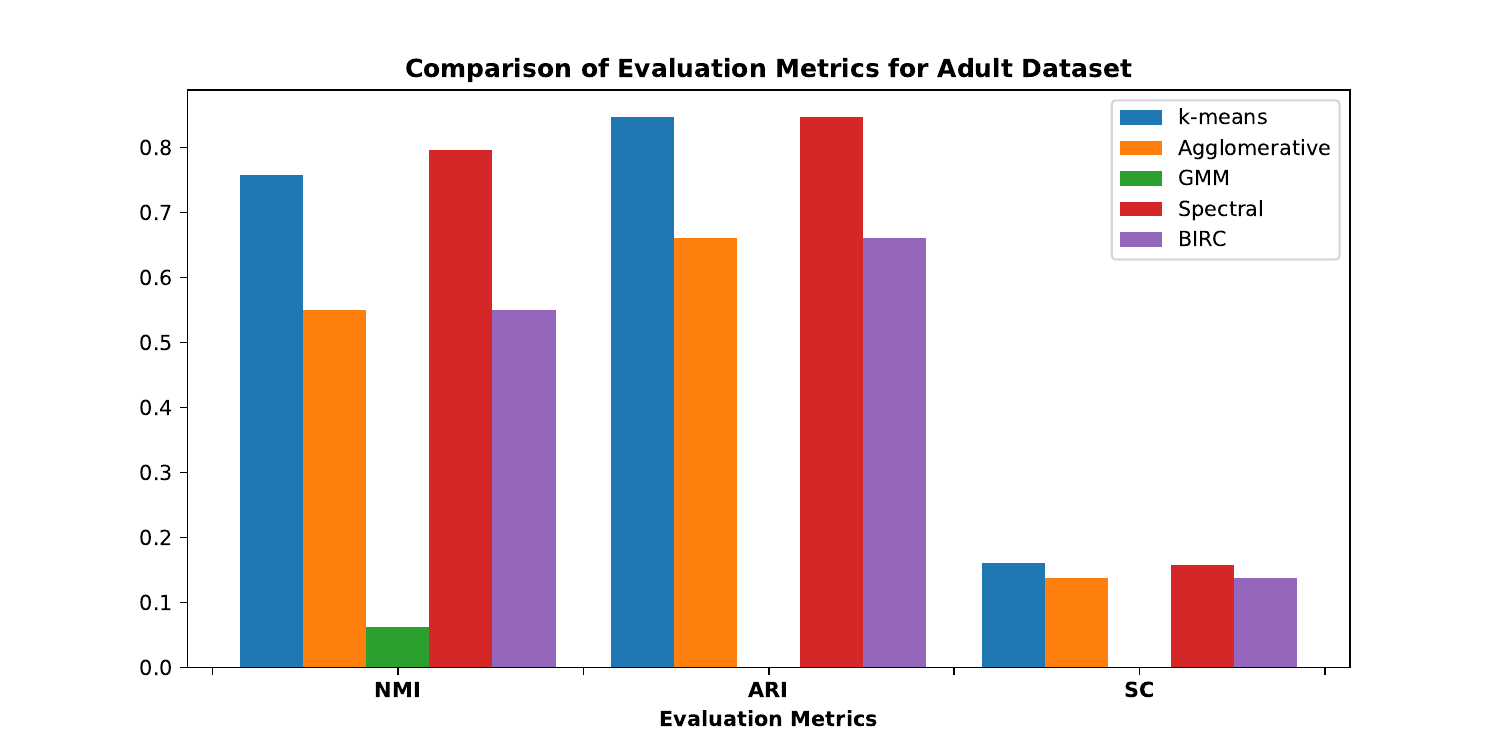}
    \includegraphics[width=.49\textwidth]{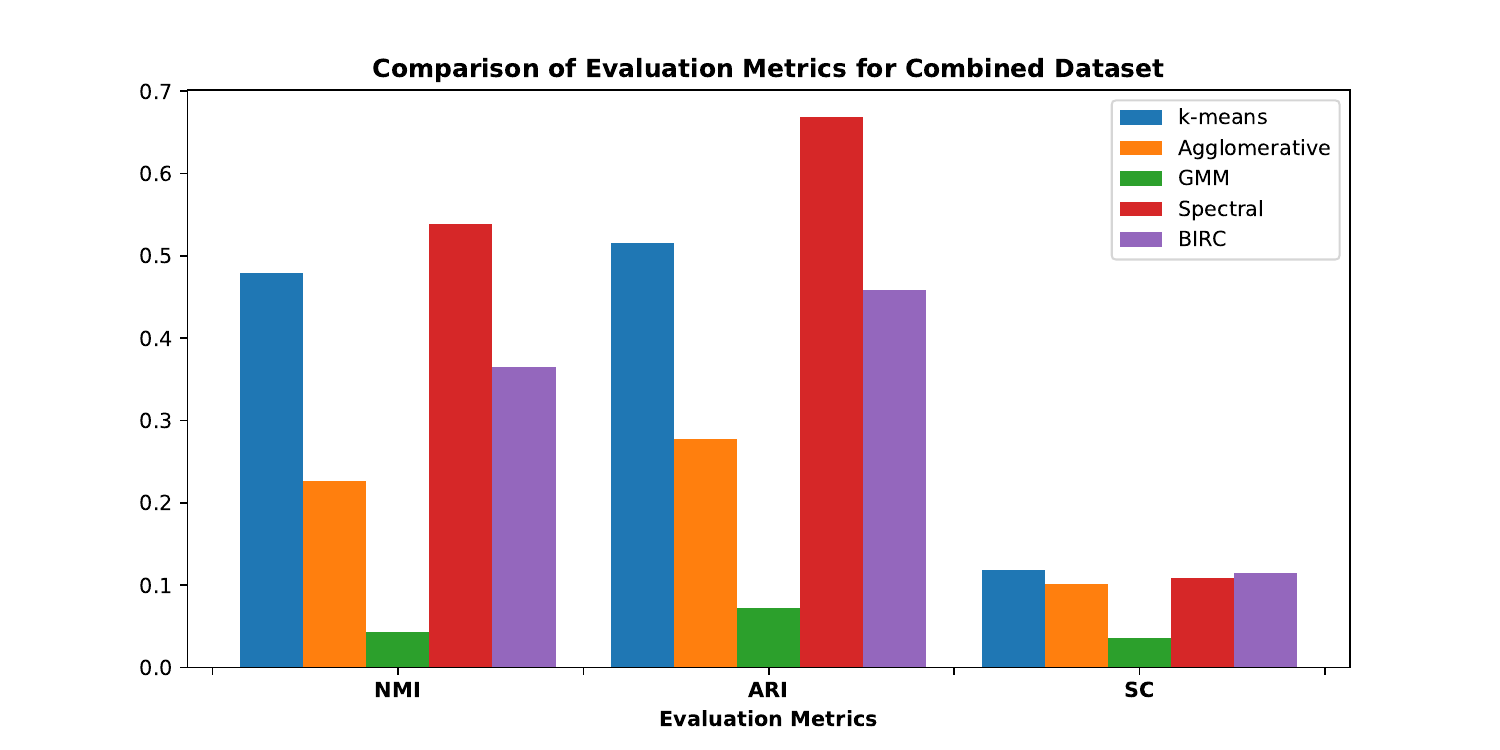}

\caption{Comparison of clustering metrics for Children, Adult and Combined dataset}
\label{fig:bar}
\end{figure*}

As almost all classification models are able to get a good accuracy of the above datasets, we become interested in observing the underlying structure of the datasets along with how machine learning models work when true labels are not provided by implementing some popular clustering algorithms on these datasets. For this purpose, we compute the Silhouette Coefficient (SC), Normalized Mutual Information (NMI), and Adjusted Rand Index (ARI) metrics to compare the selected models. Table \ref{tab:clustering} indicates the clustering results for all of the five popular clustering algorithms for the ASD Detection children, adult and combined dataset. From the table for children dataset, we can see, k-means is working well for NMI and ARI matrices and spectral clustering is good for SC metric which indicates the underlying structure of this dataset may not be in high complexity, also the dataset is not so large. For Adult and Combined datasets, we see spectral clustering is working well for both NMI \& ARI matrices and is also relatively close to the optimum SC that is found by k-means. This indicates that the underlying structure of these two datasets is a bit complex and non-linear and thus, spectral clustering is taking advantage. By using the tabular value, we generated a bar chart comparison also for visualizing the entire table result and also for easy comparison and understanding that is shown in Figure \ref{fig:bar}.

\subsection{Graphical User Interface (GUI)} 
\begin{figure*}[]
\centering

    \includegraphics[width=1.0\textwidth]{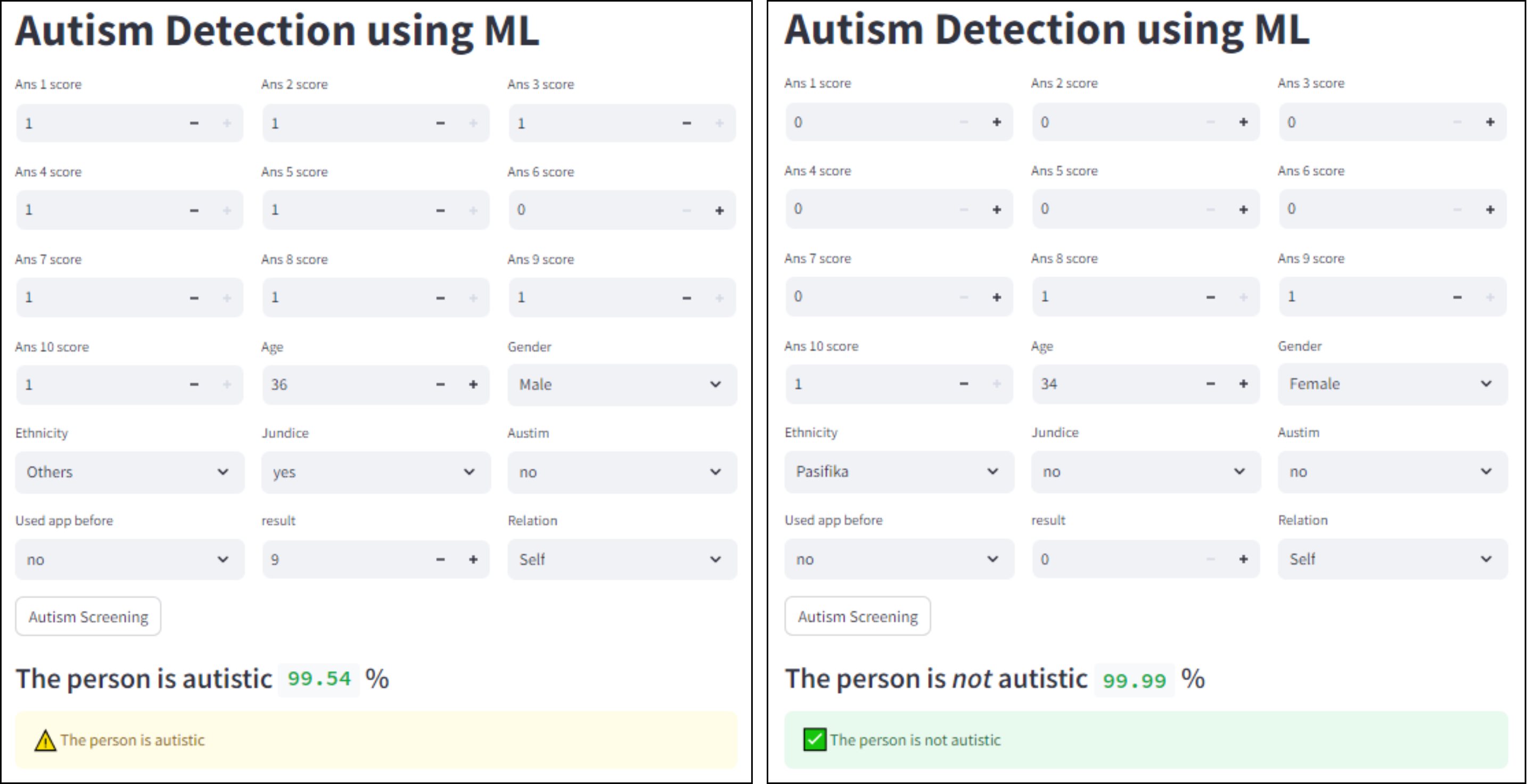}\

\caption{Graphical User Interface (GUI) for the ASD detection system}
\label{fig:gui}
\end{figure*}

Since GUIs make it easier for users to interact with, manage, and understand data or machine learning models, they have become indispensable in the fields of data science, machine learning, and artificial intelligence. Using the $streamlit$ Python library, we created a Graphical User Interface (GUI) for this study. If the machine learning models are combined with a graphical user interface (GUI), they will be more successful in identifying and screening medical diseases. The illness can then be quickly and effectively identified by a professional. To this end, we have also created a graphical user interface (GUI) that takes in an individual's traits and determines whether or not they have ASD, as well as indicating the proportion of ASD and non-ASD. In Figure \ref{fig:gui}, the implemented GUI is displayed.

After exploring all the ASD datasets with eight different learning algorithms, we have arrived at the conclusion that all of our models work well with the data. We have used eight different metrics to measure the performance of the models for the classification tasks and three different metrics for the clustering tasks. Among all the performances, we want to say that for the children dataset, the SVM and LR models are given the highest accuracy of 100\%, and for the adult dataset, the LR model is given the highest accuracy of 97.14\%. Our proposed ANN model is provided with the highest accuracy of 94.24\% for the new combined dataset, and it seems like all of the metrics indicated an almost perfect classification of the ASD cases.

\begin{table*}[t]
\caption{NMI, ARI \& SC score comparison of five different clustering algorithms for ASD Detection Children, Adult and Combined dataset. ({\em bold values are better}).}
\label{tab:clustering}
\begin{center}
{\footnotesize
\begin{tabular}{cccccccccc}
\hline
\multirow{2}{*}{\textbf{Model Name}} & \multicolumn{3}{c}{\textbf{Child Dataset}}                                                & \multicolumn{3}{c}{\textbf{Adult Dataset}}                                                & \multicolumn{3}{c}{\textbf{Combined Dataset}}                                             \\ \cline{2-10} 
                                     & \multicolumn{1}{c}{\textbf{NMI}}   & \multicolumn{1}{c}{\textbf{ARI}}   & \textbf{SC}    & \multicolumn{1}{c}{\textbf{NMI}}   & \multicolumn{1}{c}{\textbf{ARI}}   & \textbf{SC}    & \multicolumn{1}{c}{\textbf{NMI}}   & \multicolumn{1}{c}{\textbf{ARI}}   & \textbf{SC}    \\ \hline
k-means                              & \multicolumn{1}{c}{\textbf{0.615}} & \multicolumn{1}{c}{\textbf{0.628}} & 0.113          & \multicolumn{1}{c}{0.758}          & \multicolumn{1}{c}{\textbf{0.846}} & \textbf{0.160} & \multicolumn{1}{c}{0.479}          & \multicolumn{1}{c}{0.516}          & \textbf{0.119} \\ 
Agglomerative                        & \multicolumn{1}{c}{0.273}          & \multicolumn{1}{c}{0340}           & 0.077          & \multicolumn{1}{c}{0.550}          & \multicolumn{1}{c}{0.660}          & 0.137          & \multicolumn{1}{c}{0.226}          & \multicolumn{1}{c}{0.277}          & 0.102          \\ 
GMM                                  & \multicolumn{1}{c}{0.165}          & \multicolumn{1}{c}{0.140}          & 0.016          & \multicolumn{1}{c}{0.063}          & \multicolumn{1}{c}{0.001}          & 0.001          & \multicolumn{1}{c}{0.043}          & \multicolumn{1}{c}{0.072}          & 0.036          \\ 
Spectral                             & \multicolumn{1}{c}{0.340}          & \multicolumn{1}{c}{0.231}          & \textbf{0.231} & \multicolumn{1}{c}{\textbf{0.796}} & \multicolumn{1}{c}{\textbf{0.846}} & 0.158          & \multicolumn{1}{c}{\textbf{0.539}} & \multicolumn{1}{c}{\textbf{0.668}} & 0.109          \\
BIRCH                                & \multicolumn{1}{c}{0.273}          & \multicolumn{1}{c}{0.340}          & 0.077          & \multicolumn{1}{c}{0.550}          & \multicolumn{1}{c}{0.660}          & 0.137          & \multicolumn{1}{c}{0.365}          & \multicolumn{1}{c}{0.458}          & 0.115          \\ \hline
\end{tabular}
}
\end{center}
\end{table*}

\section{Conclusion}
In this paper, we explore the effectiveness of eight machine learning classification models with eight performance metrics and five clustering models with three metrics using datasets associated with Autism Spectrum Disorder (ASD) in children, adults, and a combined dataset. We find that, after accomplishing the careful hyperparameter tuning approach, all classification models have achieved good performance. Among all the performance metrics, for the children dataset SVM and Logistic Regression performed significantly and provided an accuracy of 100\% and for the adult dataset LR model produced an accuracy of 97.14\% and the ANN model showed the best performance for the combined dataset with an accuracy of 94.28\% to identify the ASD. From a clustering perspective, spectral clustering outperforms the other models. Furthermore, we identified the most significant characteristics associated with ASD and found that the A9 score, indicative of a person's aversion to physical contact, is a prominent contributing factor across adult and combined datasets while the A4 score, reflecting the challenge of comprehending others' emotions, emerges as the primary trait among children. However, it is important to acknowledge that our analysis had limitations because of the relatively limited size of the datasets associated with Autism Spectrum Disorder (ASD). To address this, we plan to apply the latest classification and clustering models to larger datasets. Additionally, we aim to extend our research by implementing deep neural network-based models capable of simultaneously learning features and classification and clustering metrics.

\section*{Author Contributions}
R.A.R.: Conceptualization, Methodology, Writing original draft, P.S.: Conceptualization, Methodology, Writing original draft, D.B.: Conceptualization, Methodology, Software, Writing - review \& editing, S.M.R.U.K.: Writing - review \& editing, I.A.: Writing - review \& editing, B.S.: Conceptualization, Methodology, Software, Writing - review \& editing.

\section*{Declaration of competing interest}
The authors declare that they have no known competing financial interests or personal relationships that could have appeared to influence the work reported in this paper.

\section*{Acknowledgments}
The authors thankfully acknowledge the organization and authors for developing the datasets and the Google Colab authority for providing the GPU services.

\bibliographystyle{cas-model2-names}

\bibliography{mybibfile}

\end{document}